\pdfoutput=1

\documentclass[11pt]{article}

\usepackage{acl} 

\usepackage{times,graphicx}
\usepackage{latexsym,booktabs}

\usepackage[T1]{fontenc}

\usepackage[utf8]{inputenc}

\usepackage{microtype}

\usepackage{inconsolata}

\usepackage{caption}
\usepackage{subcaption}
\usepackage{amsmath,wasysym,amsfonts}
\usepackage[export]{adjustbox}
\usepackage{CJKutf8}
\usepackage{multirow}

\DeclareMathOperator*{\argmin}{arg\,min}

\title{Concept Space Alignment in Multilingual LLMs}
 
\author{Qiwei Peng \and Anders S{\o}gaard \\
        University of Copenhagen\\ Denmark \\
        \texttt{\{qipe, soegaard\}@di.ku.dk}}

\begin{document}
\maketitle
\begin{abstract}

Multilingual large language models (LLMs) seem to generalize somewhat across languages. We hypothesize this is a result of implicit vector space alignment. Evaluating such alignment, we see that larger models exhibit {\em very} high-quality linear alignments between corresponding concepts in different languages. Our experiments show that multilingual LLMs suffer from two familiar weaknesses: generalization works best for languages with similar typology, and for abstract concepts. For some models, e.g., the Llama-2 family of models, prompt-based embeddings align better than word embeddings, but the projections are less linear -- an observation that holds across almost all model families, indicating that some of the implicitly learned alignments are broken somewhat by prompt-based methods.    

\end{abstract}

\section{Introduction}
Cross-lingual word embeddings are typically induced by supervised or unsupervised alignment of the word vector spaces of monolingual language models. Compression in multilingual models, i.e., parameter efficiency, can also drive {\em implicit} alignment \cite{devlin-etal-2019-bert,pires-etal-2019-multilingual,conneau-etal-2020-unsupervised}, but until recently, the mappings could still be much improved by supervised or unsupervised alignment \cite{hu-etal-2021-explicit,pan-etal-2021-multilingual}. Multilingual large language models (LLMs) are increasingly used for different tasks and demonstrate impressive ability in understanding different languages, but it is unclear whether this is a result of improved, implicit alignment, or of something else, e.g., linguistic overlap or semi-parallel subsets of training data. 

\begin{figure}[h]
\centering
\includegraphics[width=0.9\columnwidth]{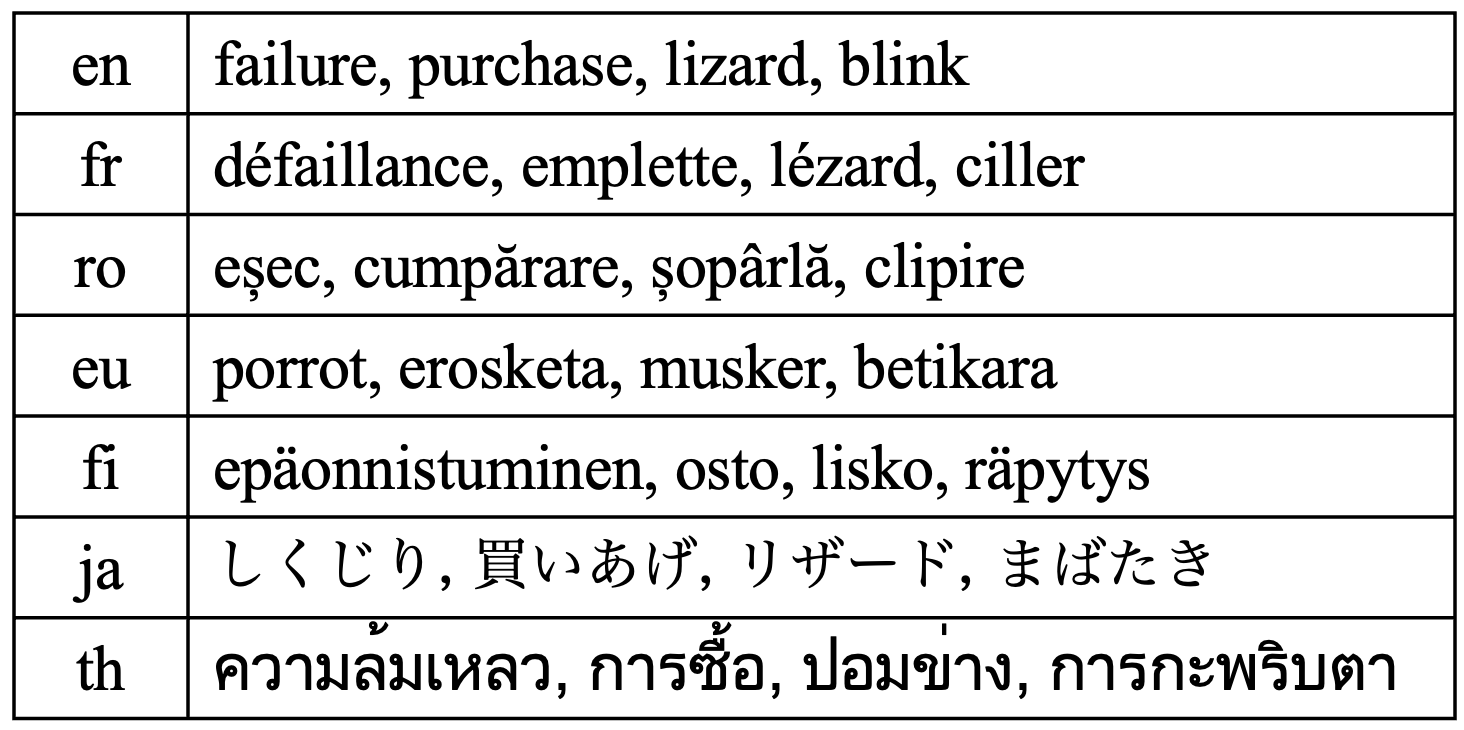}
\caption{Examples of four parallel WordNet concepts, aligned across 7 languages.}
\label{fig:data_example}
\end{figure}  

LLMs have shown promising capability to comprehend \textit{English} concepts \citep{Liao2023ConceptUI,xu2024tip}. 
Our paper sets out to evaluate concept alignment in multilingual LLMs. We aim to investigate two things: First, is there a linear mapping between corresponding concepts in different languages? Second, how does a learned linear mapping generalize to new concepts? We explore both questions by revisiting a set of techniques used in early work on bilingual dictionary induction \cite{kementchedjhieva-etal-2018-generalizing,ruder-etal-2018-discriminative,sogaard-etal-2018-limitations,kementchedjhieva-etal-2019-lost}. We evaluate multilingual LLMs {\em as if} they were bilingual dictionary induction algorithms by doing nearest neighbor search -- with cross-domain local scaling \cite{lample2018word} -- and evaluating their retrieval precision (precision@k). We first derive concept embedding in their standard way (last token or average). Since many of these models were {\em instruction fine-tuned}, we also compare {\em prompt-based embeddings} to standard techniques based on (low-level) word embeddings. We then compare their precision to retrieval rates after {\em explicit}~concept space alignment. We perform analyses with and without leakage, across multiple languages, and across both abstract and physical concepts. 

\paragraph{Contributions} Our findings across experiments with 10 LLMs and six languages suggest that linear alignment can be induced in multilingual LLMs (if sufficiently big) to map concepts across different languages. Compared to vanilla embeddings, prompt-based concept embeddings exhibit significantly lower linearity, and the gaps between before and after alignment are larger for prompt-based embeddings. This suggests that some of the implicitly learned concept alignments are broken by prompt-based methods. Prompt-based embeddings, which are now commonly used in different retrieval scenarios, seem to be less effective in extracting cross-lingually alignable embeddings, compared to vanilla embeddings. Results are generally good, but the old problem of generalization across typological distance \cite{singh-etal-2019-bert} rears its ugly face again, with Basque, Finnish, Japanese and Thai exhibiting generally lower overall performance for both experimental set-ups. Furthermore, abstract concepts exhibit better alignment than physical concepts. We suspect that it is because abstract concepts are more frequent and occur in more diverse contexts. 

\section{Experiments}
\paragraph{Concepts} We collect English noun synsets from WordNet \citep{miller1995wordnet}. For each synset, its first (most frequent) lemma name is used as the surface form of the corresponding concept. We use WordNet's hierarchical structure to filter out top-level concepts (top-5 levels) to avoid too general concepts. WordNets in other languages, such as French WordNet \citep{sagot2008building}, Basque WordNet \citep{gonzalez2012multilingual}, or Romanian WordNet \citep{dumitrescu2018rowordnet}, have similar structure and were all aligned in the Open Multilingual WordNet project (OMW) \citep{bond2016cili}. To produce a repository of parallel semantic concepts, we collect synsets with shared ID across different WordNets, after removing duplicate concepts. In total, we obtain 4,397 parallel concepts across 7 different languages (English, French, Romanian, Basque, Finnish \citep{linden2014possible}, Japanese \citep{bond2009enhancing} and Thai \citep{thoongsup2009thai}); had we included more languages, the number of parallel concepts would have been prohibitively small. The 4,397 concepts were divided into abstract (e.g., happiness) and physical (e.g., vehicle) concepts. See Table \ref{tab:data_statistics} for data characteristics, and Figure \ref{fig:data_example} for examples of parallel concepts.

\begin{table}[h!]
\centering
\resizebox{0.7\columnwidth}{!}{%
\begin{tabular}{|l|c|c|c|}
\hline
      & Abstract & Physical & Total \\ \hline
Train & 1500     & 1500     & 3000  \\ \hline
Test  & 419      & 978      & 1397  \\ \hline
Total & 1919     & 2478     & 4397  \\ \hline
\end{tabular}%
}
\caption{The statistics of the parallel concept dataset. We use 1000, 2000, or 3000 concepts for training.}
\label{tab:data_statistics} 
\end{table}   

\noindent To create a \textbf{seed dictionary} (training data) for supervised alignment, we randomly sample 3,000 parallel concepts,\footnote{See Appendix for results with 1,000 or 2,000 concepts.} including 1,500 abstract concepts and 1,500 physical concepts. The 3,000 concepts are used to induce the linear mapping.  

\paragraph{LLMs} We experiment with four different LLM families with varying sizes: Llama2 (7B, 13B, 70B) \citep{touvron2023llama}, mT0 (1.2B, 3.7B, 13B), BLOOMZ (1B7, 3B, 7B1) \citep{muennighoff2022crosslingual}, and Aya101 (13B) \citep{ustun2024aya}. We use \textit{two different} concept space extraction methods (vanilla and prompt-based). The vanilla method simply uses the last token representation as the concept embedding for decoder-only models (Llama2 and BLOOMZ); and the average embedding of the last hidden layer of the encoder as the concept embedding for encoder-decoder models (mT0 and Aya101)\footnote{This is decided by preliminary experiment results.}. The prompt-based extraction method exploits the fact that all these models were instruction-tuned. The template we use for prompt-based extraction is adapted from \citet{li2023angle} and shown as follows: 

{\footnotesize
\begin{itemize}
    \item[] {\sf Summarize concept [text] in one [lang] word:}
\end{itemize}
}

\noindent where {\sf [text]} and {\sf [lang]} will be replaced by the corresponding concept (in the source language) and the language name (in adjectival form), e.g.,"summarize concept "\begin{CJK}{UTF8}{min}動物\end{CJK}" in one Japanese word" for the concept \textit{animal}\footnote{The [text] of a concept can be made up of multiple words.}. The prompt-based concept embedding is that of the last hidden state.

\begin{figure*}[h!]
\vspace{-5pt}
     \centering
     \begin{adjustbox}{minipage=\textwidth,scale=0.9} 
     \begin{subfigure}[t]{0.48\textwidth}
         \centering
         \includegraphics[width=\textwidth]{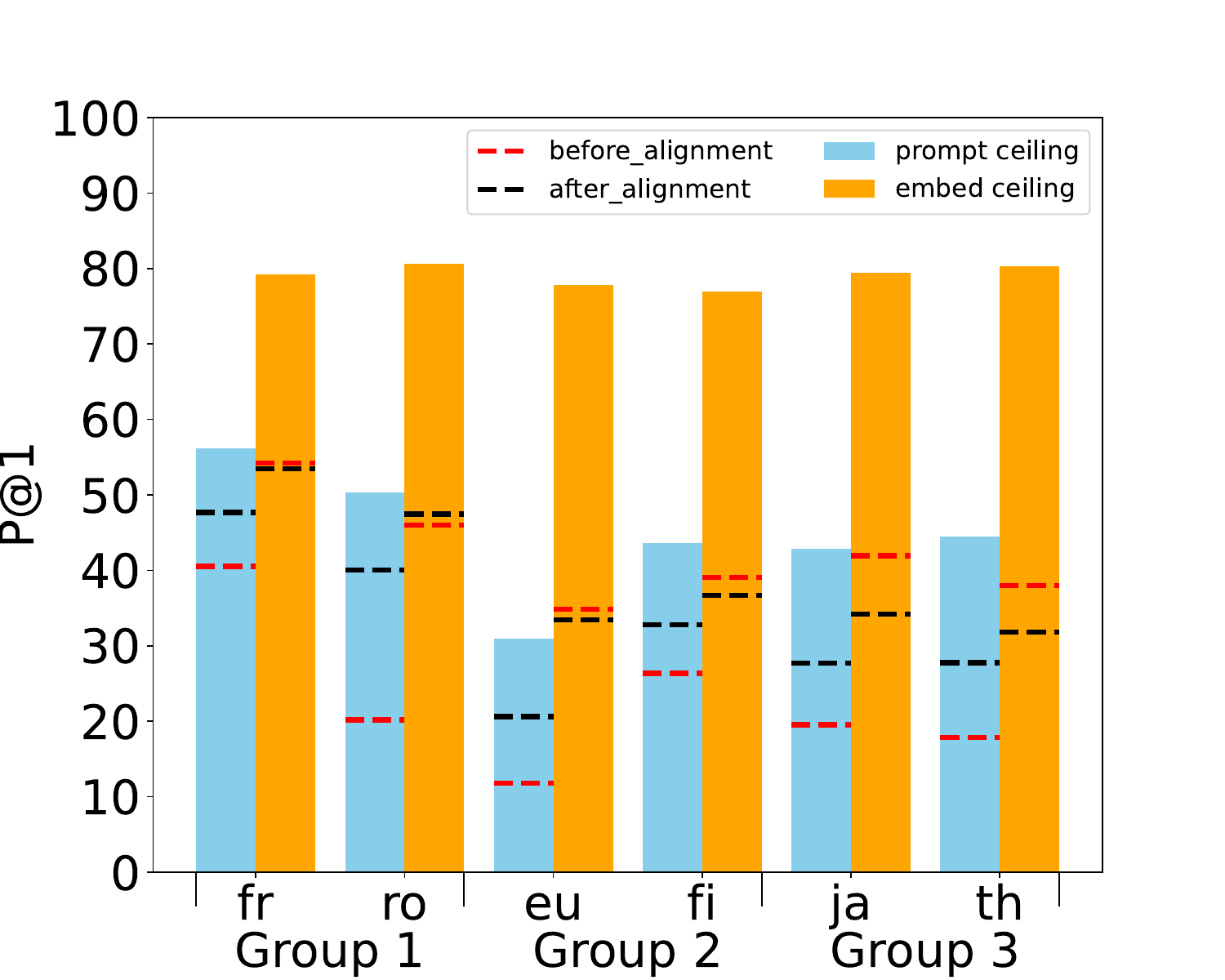}
         \caption{Aya101 (13B)}
    \end{subfigure}
    \hfill
    \begin{subfigure}[t]{0.48\textwidth}
         \centering
         \includegraphics[width=\textwidth]{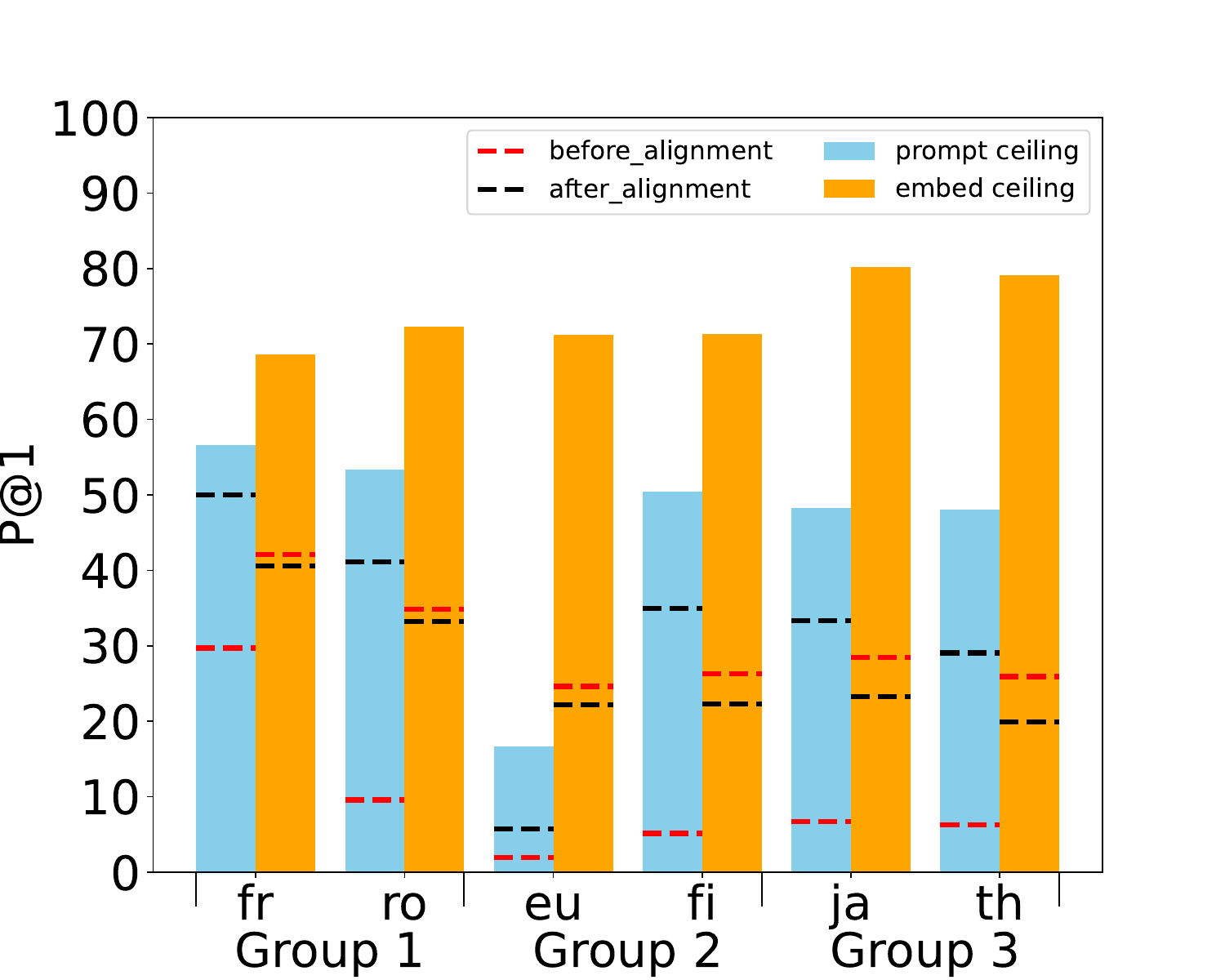}
         \caption{mT0-xxl (13B)}
    \end{subfigure}
    \newline
     \begin{subfigure}[t]{0.48\textwidth}
         \centering
         \includegraphics[width=\textwidth]{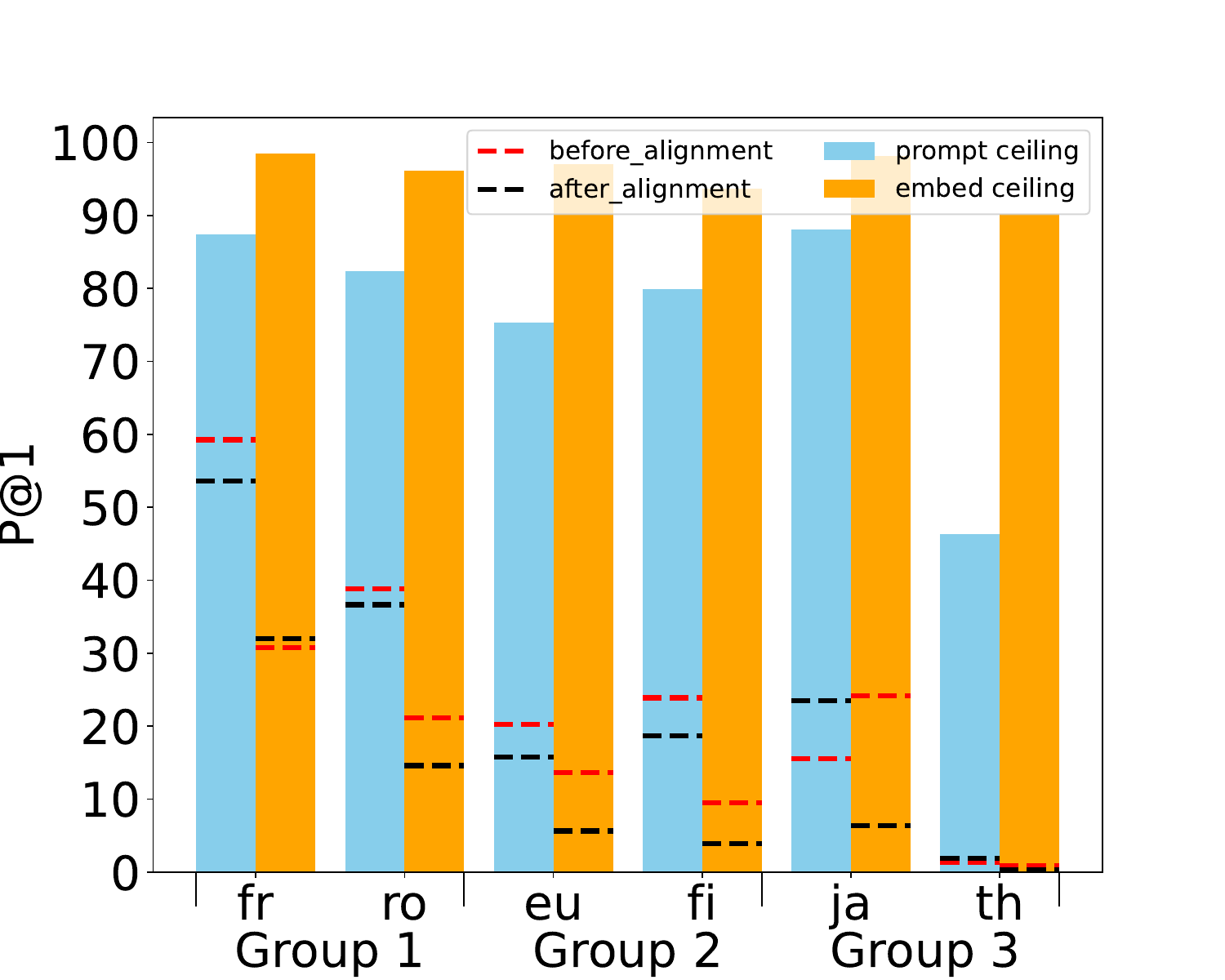}
         \caption{Llama2-13B}
     \end{subfigure}
     \hfill
     \begin{subfigure}[t]{0.48\textwidth}
         \centering
         \includegraphics[width=\textwidth]{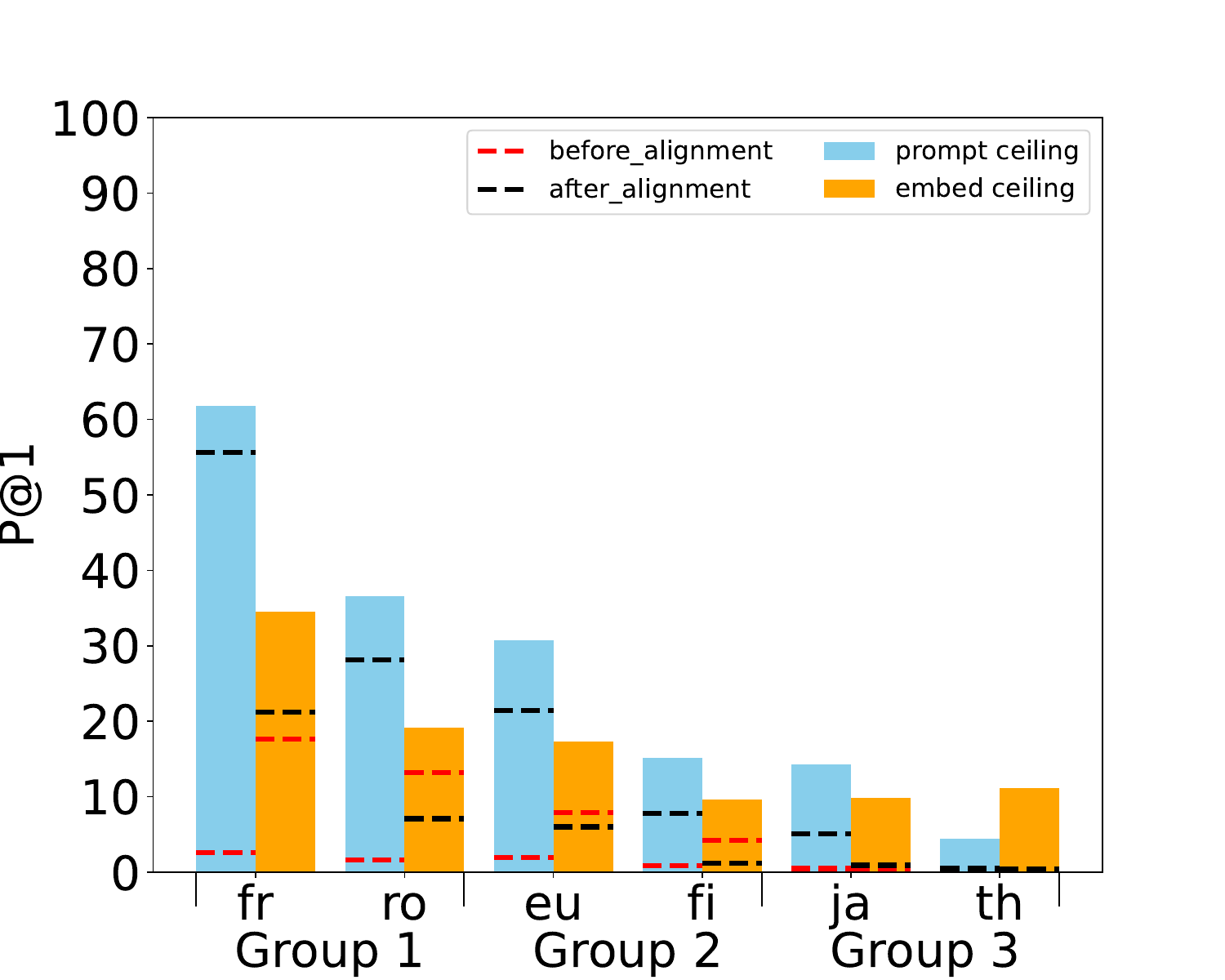}
         \caption{BLOOMZ-7B1} 
     \end{subfigure}
\end{adjustbox}
\caption{Performance (P@1) of different LLMs on the concept alignment evaluation when using a seed dictionary of 3,000 concepts. X-axis: Languages, we further divide these languages into three groups, where \textbf{Group 1} is Indo-European, \textbf{Group 2} includes languages that are not Indo-European but still in Latin script, while \textbf{Group 3} refers to languages that are not Indo-European and not in Latin script. Y-axis: We report Precision@1.}
\label{fig:main_exp}
\end{figure*} 

\paragraph{Alignment and Retrieval} We rely on Procrustes Analysis \citep{schonemann1966generalized}, a form of statistical shape analysis, to discover good linear transformations (e.g., translation, rotation, and scaling) between concept spaces in different languages. Suppose $X$ and $Y$ are two matrices of size $n \times d$ ($n$ is the seed dictionary size, $d$ is the embedding size) such that the $i$th row in $X$ is an embedding of concept $c_i$ in one language, and the $i$th row in $Y$ is $c_i$'s embedding in the other language. The linear transformation is derived through singular value decomposition (SVD) of $YX^{T}$: 
\begin{equation}
    W^{*} = \argmin_{W \in O_{d(\mathbb{R})}}||WX - Y||_{F} = UV^{T}
\end{equation}
where $U \Sigma V^{T} = \text{SVD}(YX^{T})$. With $W^{*}$, we transform source language concept embeddings $X$ into the \textit{English} (target) vector space. We then perform cross-domain local scaling (CSLS) to retrieve the most similar concepts.\footnote{We also ran experiments with vanilla nearest neighbor search as our retrieval method, but CSLS outperforms nearest neighbor search by some margin. So, we report results with CSLS.} We use precision@k (P@k) as our performance metric. 

\paragraph{Main Results} We present the main results\footnote{Full results with all model sizes, training sizes, and different $k$-values for P@k are presented in the Appendix.} in Figure \ref{fig:main_exp}. For each model, we report three results: 1) the {\em upper bound} (leaky) on performance for supervised linear alignment, using the train seed {\em and} the test seed for inducing the dictionary (orange/blue bar), which we refer to as the {\em ceiling} and reveals to what extent there exists a linear mapping; 2) {\em before-align} performance (red dashed line), retrieval bilingual concept pairs directly from the {\em raw} LLM (vanilla word or prompt) embeddings; 3) {\em after-align} performance (black dashed line), which is the performance of non-leaky, supervised mapping (using 3,000 concepts as the seed dictionary) into the English vector space, with CSLS as our retrieval method. Orange bars indicate vanilla word embedding strategy (last-token, or average embedding), while blue bars refer to results for prompt-based embedding.

All multilingual LLMs (except BLOOMZ) can induce good concept alignments, as indicated by the upper bound performance. In general, within the same model family, a larger model size leads to better alignment. The ceiling is highest for vanilla word embeddings in Llama2-13B, indicating near-isomorphisms between monolingual concept spaces at this level. The prompt-based embeddings are less linear, indicating that partial isomorphisms induced prior to prompting are corrupted. For after-align performance, we generally see the highest performance for Indo-European languages (Group 1) and the lowest for non-Indo-European languages with non-Latin scripts (Group 3). Similarly, a larger model size and larger seed dictionary generally improve the concept alignment. On Group 2 and 3, mT0 and Aya101 show better before-align performance compared to other models. In some cases, results are extremely good. Llama2-13B with prompt-based embeddings exhibits a P@1 score of 59.27\% before alignment for French, for example. This means that the model has induced perfect alignment of 3/5 concepts in the absence of any explicit supervision. It is interesting to see the gap between the red and black dashed lines. The size of this gap indicates how much of the (alignable part of the) concept space was {\em not} aligned, with given seed dictionary. For vanilla word embeddings, the gaps are relatively small, but for prompt-based embeddings the gaps tend to be much larger, again indicating that prompting somewhat breaks the implicitly learned concept alignment.  

\paragraph{Abstract vs. Physical} 
We analyze performance differences across abstract and physical concepts. To make a fair comparison, we randomly down-sample\footnote{See Appendix for numbers without down-sampling.} physical concepts and compare retrieval performance across the two classes. In this section, we report P@1 with models that have comparable model sizes (7B/13B) in each family; results for the other models can be found in the Appendix. As shown in Table \ref{tab:abs_phy_exp}, all models generally have better alignment performance on abstract concepts compared to physical concepts.  

\begin{table}[h!]
\centering
\resizebox{1\columnwidth}{!}{%
\renewcommand{\arraystretch}{1.3}
\begin{tabular}{|c|l|c|c|c|c|c|c|}
\hline
                            & \textbf{} & fr    & ro    & eu    & fi    & ja    & th    \\ \hline
\multirow{2}{*}{Llama2-13B} & Abstract  & \textbf{63.48} & \textbf{46.06} & \textbf{17.42} & \textbf{21.00} & \textbf{26.01} & \textbf{2.15} \\  
                            & Physical  & 50.12 & 33.41 & 14.08 & 18.62 & 23.39 & 1.91 \\ \hline
\multirow{2}{*}{BLOOMZ-7B1} & Abstract  & \textbf{64.92} & \textbf{33.41} & \textbf{27.92} & \textbf{10.74} & \textbf{10.26} & \textbf{1.19}  \\  
                            & Physical  & 52.51 & 27.45 & 18.62 & 6.44 & 4.53 & 0.00  \\ \hline
\multirow{2}{*}{mT0-xxl (13B)}    & Abstract  & \textbf{59.90} & \textbf{49.88} & \textbf{7.88} & \textbf{38.90} & \textbf{38.42} & \textbf{34.37} \\  
                            & Physical  & 46.78 & 41.29 & 5.73 & 37.23 & 36.28 & 28.64 \\ \hline
\multirow{2}{*}{Aya101 (13B)}     & Abstract  & \textbf{58.47} & \textbf{52.27} & \textbf{27.68} & \textbf{40.81} & \textbf{36.28} & \textbf{32.70} \\  
                            & Physical  & 44.63 & 36.75 & 18.38 & 30.79 & 26.73 & 29.12 \\ \hline
\end{tabular}%
}
\caption{The results (P@1) for abstract and physical concepts. We report after-align results for prompt-based embedding and comparable sizes (13B/7B) of each model family.} 
\label{tab:abs_phy_exp} 
\end{table}    

What explains this very consistent finding? One hypothesis would be that physical nouns are more ambiguous, since they often source metaphor and metonymy. However, our words for abstract concepts have more senses in WordNet (2.94) than our words for physical concepts (1.96); see Table \ref{tab:concept_sense_freq}. Instead, we found another, simpler explanation. Frequency statistics (obtained from the English Wikipedia dump of 2023-04-13) relevant that the abstract concept words are considerably more frequent than the physical concept words, which makes sense, as abstract concepts apply very generally across contexts and domains.

\begin{table}[h!]
\centering
\resizebox{0.8\columnwidth}{!}{%
\begin{tabular}{|l|c|c|}
\hline
                    & Abstract & Physical \\ \hline
avg \# of senses     & 2.94     & 1.96     \\ \hline
median \# of senses  & 2        & 1        \\ \hline
avg \# of counts & 103,934   & 28,762    \\ \hline
median \# of count & 12,787    & 5,122     \\ \hline
\end{tabular}%
}
\caption{Number of senses and frequency of words.}
\label{tab:concept_sense_freq}
\end{table}

\section{Discussion and Related Work}
\paragraph{Related Work}
The idea that distributional representations facilitate cross-lingual alignment goes back to explicit semantic analysis \cite{gabrilovich2007computing}, but the idea of training multilingual, neural language models also has a long history. Such models have traditionally used explicit alignment objectives, e.g., either from word alignments, bilingual dictionary seeds \cite{lample2018word, li2024prealign}, or by training on mixed corpora constructed using such resources \cite{gouws-sogaard-2015-simple, workshop2022bloom, chai2024xcot}. Cross-lingual generalization has been studied in different NLP tasks, including question answering \citep{artetxe2020cross}, commonsense reasoning \citep{ponti2020xcopa, lin2022few}, code generation \citep{peng2024humaneval}, and knowledge transfer and consistency \citep{xu-etal-2023-language-representation, qi-etal-2023-cross}. Cross-lingual word alignment also has a long history by examining bilingual lexicon induction \citep{xing-etal-2015-normalized, sogaard-etal-2018-limitations, li-etal-2023-bilingual}. For concept understanding specifically, previous works have examined concept understanding in LLMs by definition matching \citep{xu2024tip}, hypernym/hyponym detection \citep{Liao2023ConceptUI,shani2023towards}, and relation discovery \citep{gu2023language}. However, they are limited to the English language only.  

\paragraph{Linear Alignment}
We saw that concepts are represented in similar ways across languages in multilingual LLMs, as shown in the upper bound. This indicates structural similarities and facilitates cross-lingual transfer. Prompt-based embeddings exhibit significantly lower linearity compared to word embeddings, and the gaps between before and after alignment are larger for prompt-based embeddings. Both things suggest that some of the implicitly learned concept alignment is broken by the prompt-based method. On the other hand, prompt-based embeddings demonstrate larger improvements with explicit post-hoc alignment while supervised alignment struggles to improve on vanilla word embeddings. 

\paragraph{Difference in Languages} The degree of isomorphism to English is similar across languages, as indicated by the upper bounds on performance. All concept spaces are (almost) equally alignable. However, the induced maps generalize much better across typologically related (Indo-European) languages: French and Romanian. Generalization is considerably poorer for the other two groups. 

\paragraph{Types of Concepts} Though previous works show that physical concepts do better than abstract ones in bilingual dictionary induction \cite{kementchedjhieva-etal-2019-lost}, as well as in related tasks such as hypernym detection \citep{Liao2023ConceptUI}, we show that abstract concepts tend to align better across different languages, as shown in Table \ref{tab:abs_phy_exp}. This, however, was explained by a spurious correlation with frequency. It would be interesting to control for frequency in future error analysis. 

\section{Conclusion}
We evaluated concept alignment on multilingual LLMs by revisiting the traditional bilingual dictionary induction task, but with semantic concepts rather than words. Our experiments show that multilingual LLMs exhibit high-quality, linear concept alignment across different languages. However, the ability of supervised maps to generalize varied across different models, languages, and ways of obtaining embeddings.  

\section*{Limitations}
Because of the small overlap between multilingual WordNets, we only include six (6) test languages. While this is too small a set of languages to draw universally applicable conclusions. Fortunately, the set includes both Indo-European and non-Indo-European languages, as well as both Latin script and non-Latin script languages. We also limited ourselves to studying nouns; for how linear alignment generalizes to other parts of speech, see \citet{kementchedjhieva-etal-2018-generalizing} and \citet{hartmann-sogaard-2018-limitations}.  

\section*{Ethical Considerations}
We do not anticipate any risks in the work. In this study, our use of existing artifacts is consistent with their intended purposes. Semantic concepts are collected from previously published and publicly available resources (WordNets). Aya101\footnote{https://huggingface.co/CohereForAI/aya-101\#model-summary}, BLOOMZ, and mT0 models have Apache-2.0 License\footnote{https://github.com/bigscience-workshop/xmtf/blob/master/README.md}. Llama2 models are under the LLAMA 2 Community License\footnote{https://ai.meta.com/llama/license/}.

\section*{Acknowledgement} 
We would like to thank all anonymous reviewers for their insightful comments and feedback. This work was supported by DisAI - Improving scientific excellence and creativity in combating disinformation with artificial intelligence and language technologies, a project funded by European Union under the Horizon Europe, GA No. 101079164.

\bibliography{emnlp2023}
\bibliographystyle{acl_natbib}

\appendix

\section{Language Resource and Shared Vocabulary}
\label{sec:appendix_1}
We report the estimated resource level for the seven languages we experimented in this work. The number, which has been widely used to indicate resource availability, is taken from the CC100 XL corpus\citet{lin2022few}. 

\begin{table}[h!]
\centering
\resizebox{0.5\columnwidth}{!}{%
\begin{tabular}{|c|c|}
\hline
Lang     & Tokens (M) \\ \hline
English  & 803,527    \\ \hline
French   & 77,420     \\ \hline
Japanese & 66,054     \\ \hline
Romanian & 24,176     \\ \hline
Finnish  & 16,804     \\ \hline
Thai     & 10,842     \\ \hline
Basque   & 105        \\ \hline
\end{tabular}%
}
\caption{The statistics of language resource level for 7 languages used in this work.}
\label{tab:language_resource}
\end{table}

One reason of the cross-lingual alignment might be that among the chosen concepts some had the same word form across languages with Latin scripts. To investigate this, we additionally calculate the ratio of identical word forms (compared to English concepts) for languages with Latin scripts:
\begin{itemize}
    \item 653 out of 4397 for French (roughly 15\%)
    \item 449 out of 4397 for Romanian (roughly 10\%) 
    \item 243 out of 4397 for Basque (roughly 5\%) 
    \item 172 out of 4397 for Finnish (roughly 4\%)
\end{itemize}
This ratio remains similar when counted only on the test split. We can observe from above that the ratio is quite limited.  

\section{Full Experimental Results}
\label{sec:appendix_2}
In the Appendix, we report our full experimental results across different models with varying model sizes, seed dictionary sizes, different k-values for P@K, in following figure and tables (Figure 3 and Table 4-23). These results provide full scope of our analysis, allowing for an in-depth comparison of model performances. 

For different models, we use their HuggingFace PyTorch implementation\footnote{https://huggingface.co/bigscience/bloomz-\{1b7,3b,7b1\},\newline https://huggingface.co/meta-llama/Llama-2-\{7,13,70\}b-chat-hf,\newline https://huggingface.co/bigscience/mt0-\{large,xl,xxl\},\newline https://huggingface.co/CohereForAI/aya-101}. For Procrustes Analysis, we utilize the MUSE\footnote{https://github.com/facebookresearch/MUSE} package. All experiments are run on a single NVIDIA A100 GPU. 

\begin{figure*}[h!]
     \centering
     \begin{subfigure}[t]{0.32\textwidth}
         \centering
         \includegraphics[width=\textwidth]{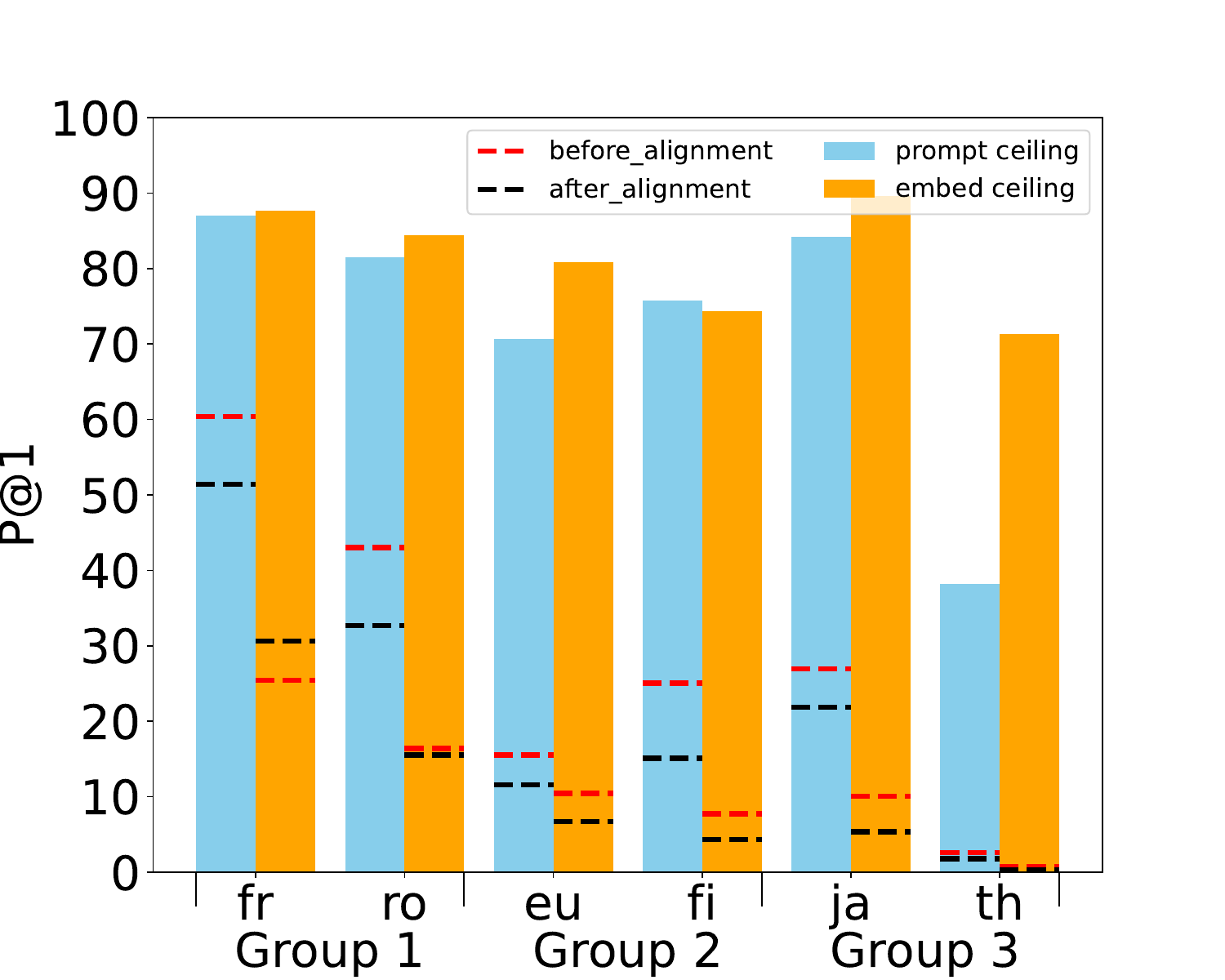}
         \caption{Llama2-7B}
     \end{subfigure}
     \hfill
     \begin{subfigure}[t]{0.32\textwidth}
         \centering
         \includegraphics[width=\textwidth]{figs/llama2_13b.pdf}
         \caption{Llama2-13B}
     \end{subfigure}
     \hfill
     \begin{subfigure}[t]{0.32\textwidth}
         \centering
         \includegraphics[width=\textwidth]{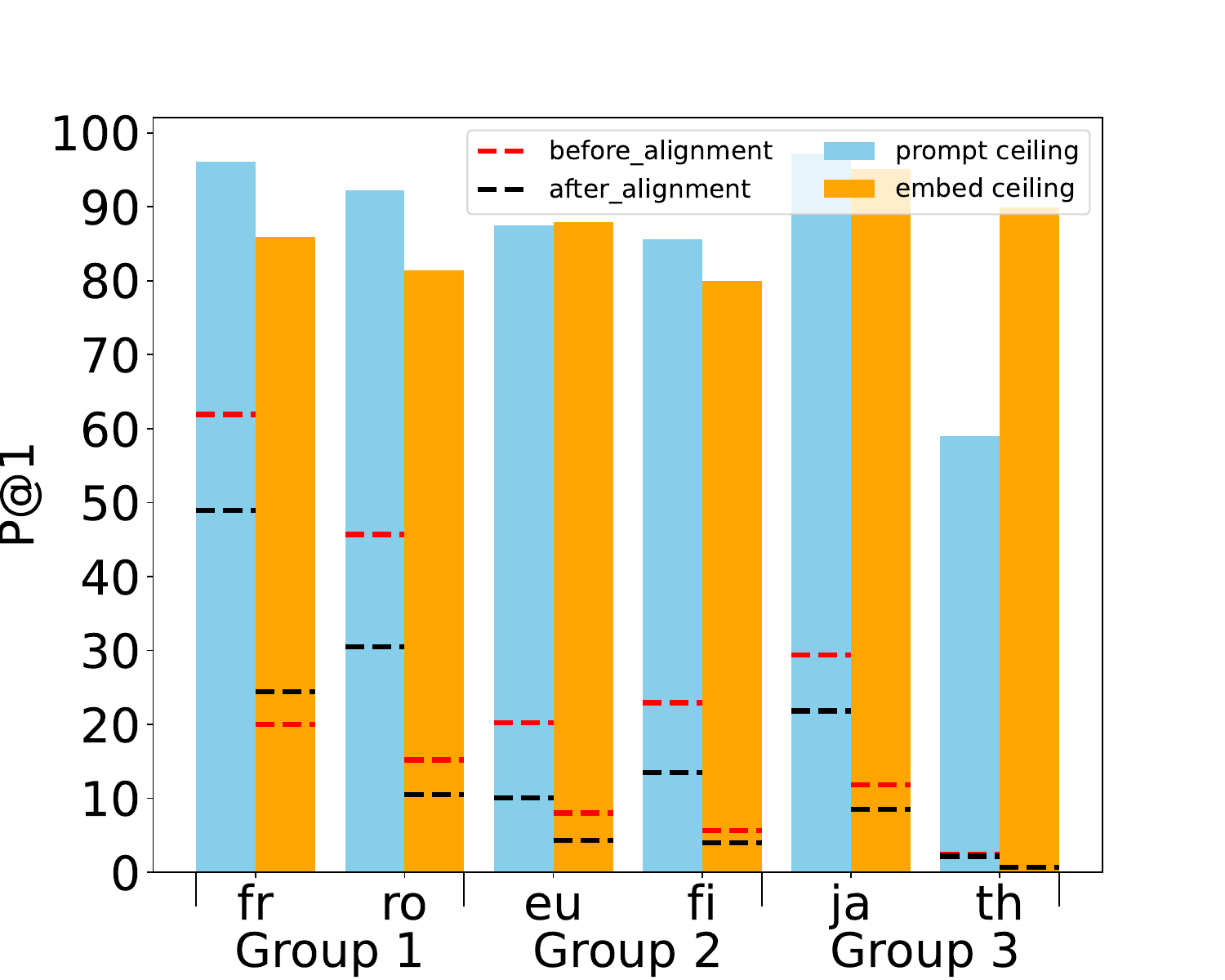}
         \caption{Llama2-70B}
     \end{subfigure}
     \newline
     \begin{subfigure}[t]{0.32\textwidth}
         \centering
         \includegraphics[width=\textwidth]{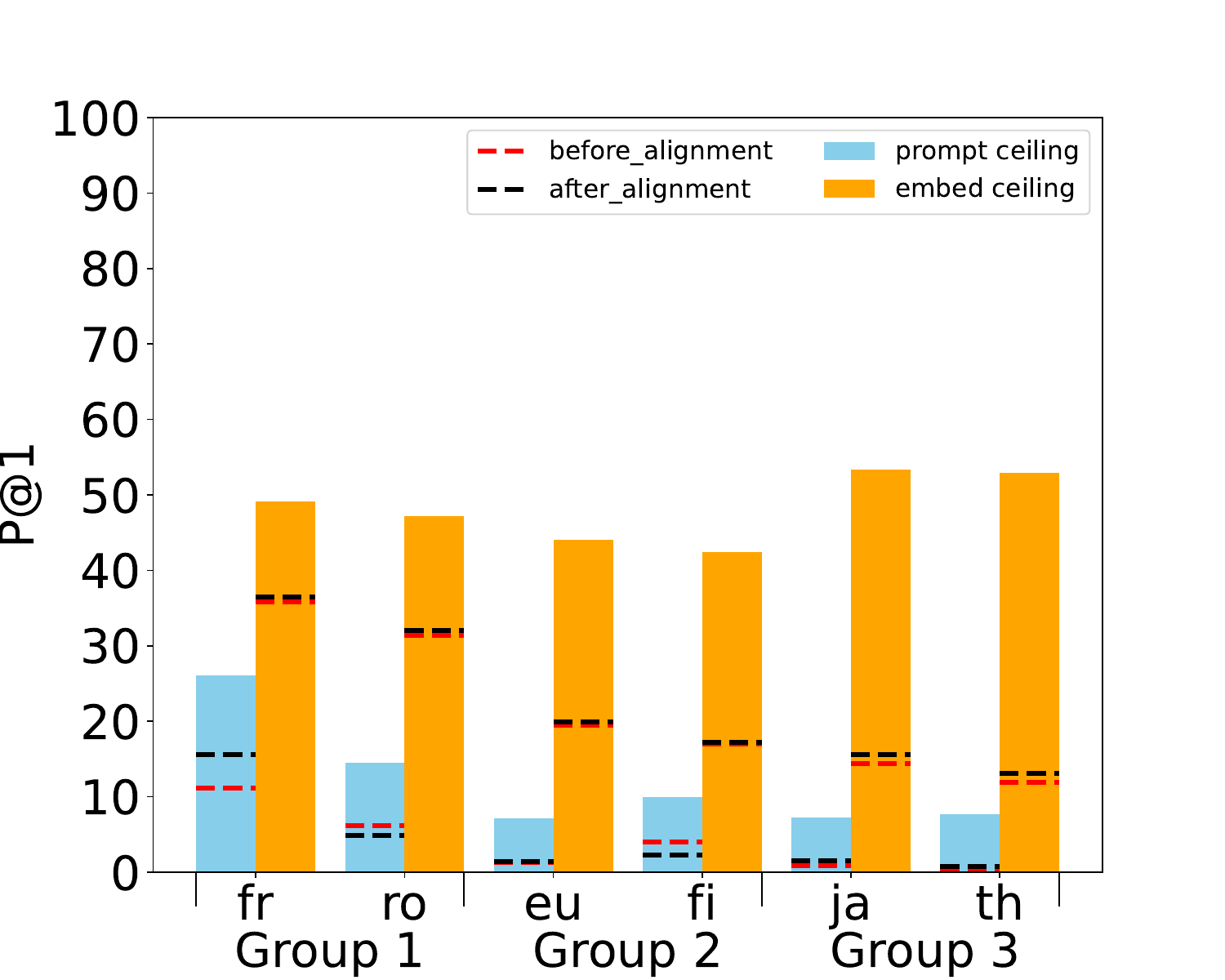}
         \caption{mT0-large (1.2B)}
     \end{subfigure}
     \hfill
     \begin{subfigure}[t]{0.32\textwidth}
         \centering
         \includegraphics[width=\textwidth]{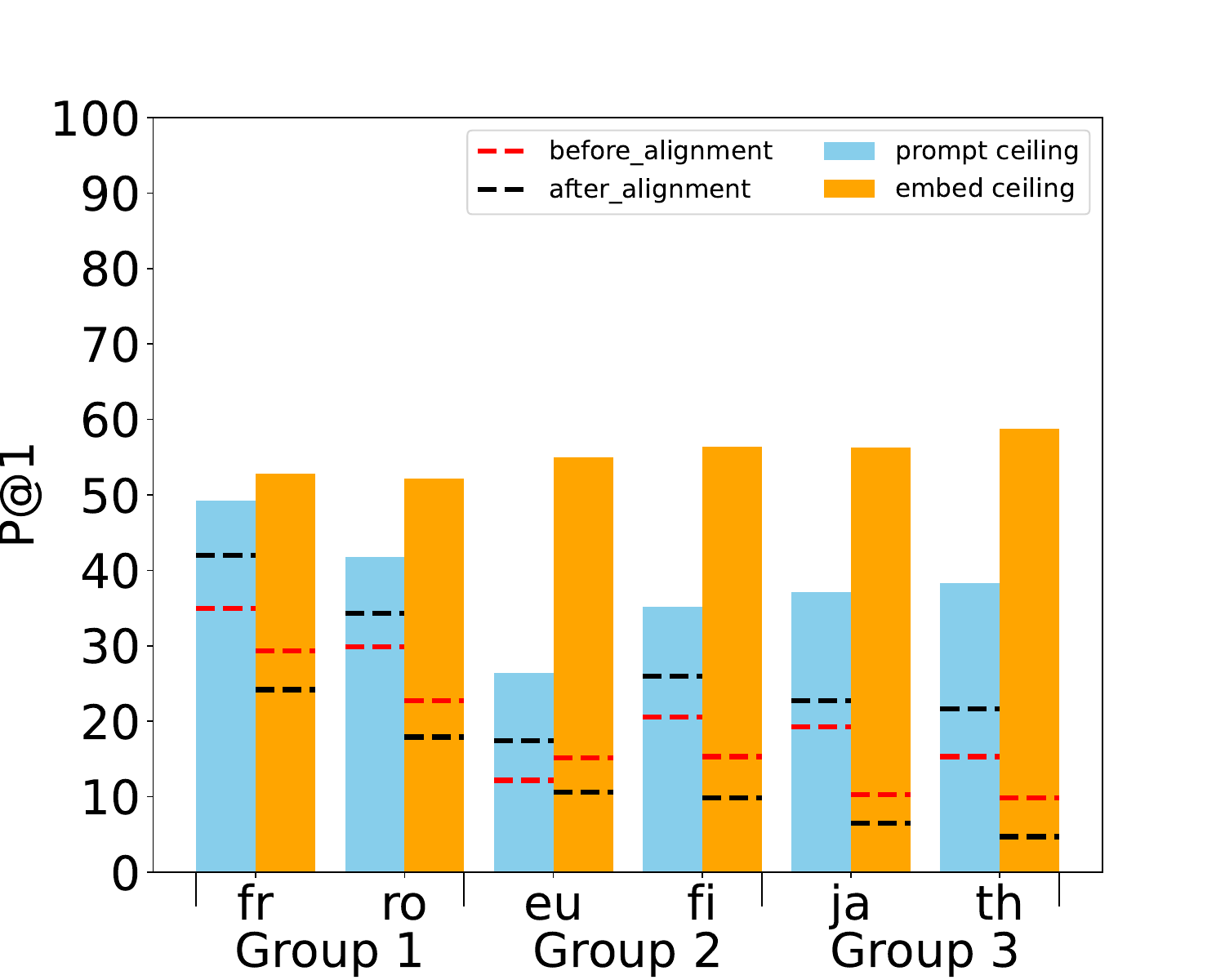}
         \caption{mT0-xl (3.7B)}
     \end{subfigure}
     \hfill
     \begin{subfigure}[t]{0.32\textwidth}
         \centering
         \includegraphics[width=\textwidth]{figs/mT0_xxl.pdf}
         \caption{mT0-xxl (13B)}
     \end{subfigure}
     \newline
     \begin{subfigure}[t]{0.32\textwidth}
         \centering
         \includegraphics[width=\textwidth]{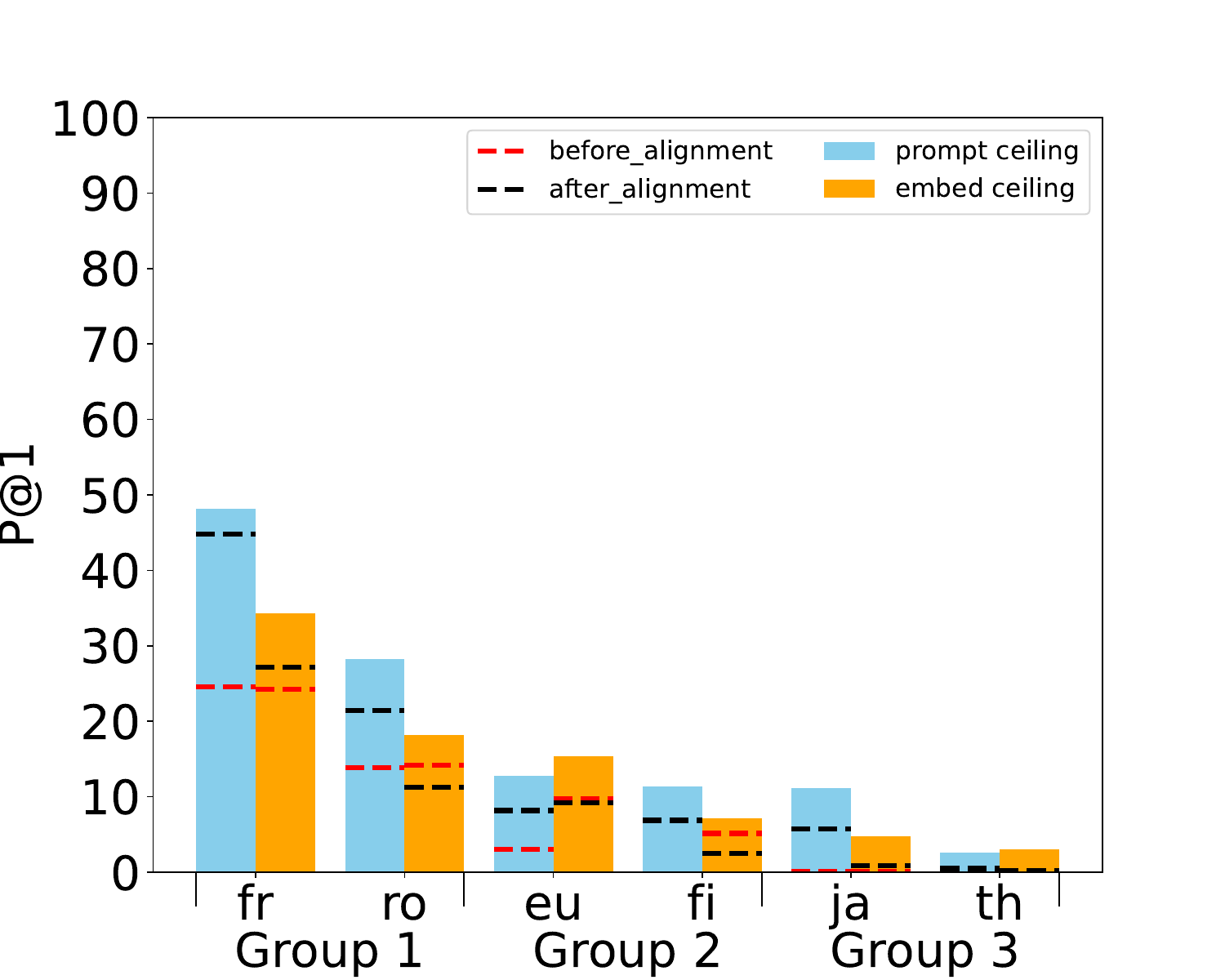}
         \caption{BLOOMZ-1B7}
     \end{subfigure}
     \hfill
     \begin{subfigure}[t]{0.32\textwidth}
         \centering
         \includegraphics[width=\textwidth]{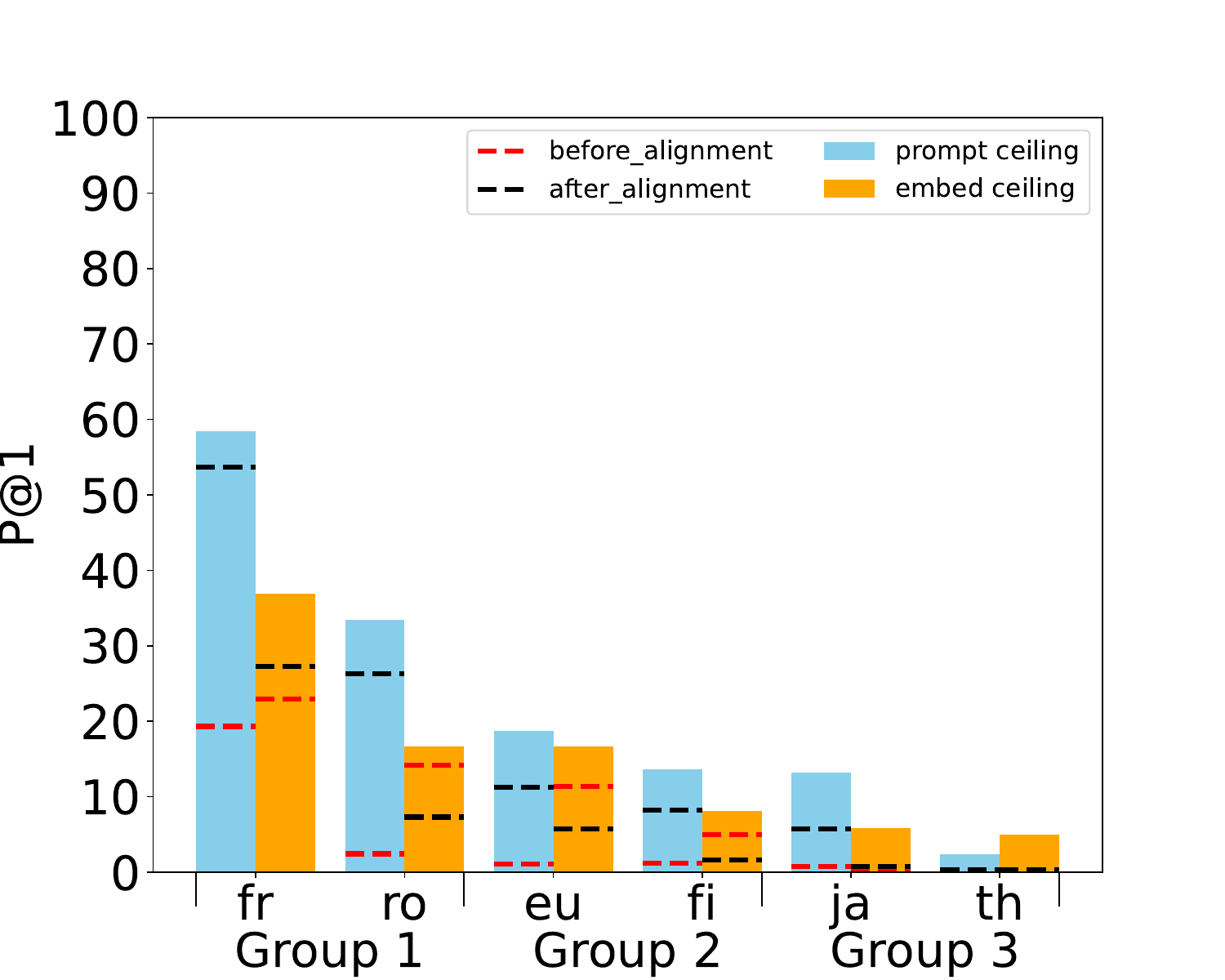}
         \caption{BLOOMZ-3B}
     \end{subfigure}
     \hfill
     \begin{subfigure}[t]{0.32\textwidth}
         \centering
         \includegraphics[width=\textwidth]{figs/BLOOMZ_7b1.pdf}
         \caption{BLOOMZ-7B1} 
     \end{subfigure}
     \newline
     \begin{subfigure}[t]{0.32\textwidth}
         \centering
         \includegraphics[width=\textwidth]{figs/aya101.pdf}
         \caption{Aya101 (13B)}
     \end{subfigure}
     \hfill
\caption{Performance (P@1) of different LLMs on the concept alignment evaluation when using a seed dictionary of 3000 pairs. X-axis: Languages, we further divide these languages into three groups, where \textbf{Group 1} is Indo-European, \textbf{Group 2} includes languages that are not Indo-European but still in Latin script, while \textbf{Group 3} refers to languages that are not Indo-European and not in Latin script. Y-axis: We report Precision@1.}
\label{fig:full_experiments}
\end{figure*}

\begin{table*}[h!]
\centering
\resizebox{0.8\textwidth}{!}{%
\begin{tabular}{|c|l|c|c|c|c|c|c|}
\hline
                    & \multicolumn{7}{c|}{\begin{tabular}[c]{@{}c@{}}EN\\ (P@K: 1000/2000/3000) \end{tabular}}                                   \\ \hline
Language            & P@K  & Before-Alignment & Ceiling Performance & Eval - Both       & Eval - Abstract   & Eval-Physical (all) & Eval - Physical (downsample)   \\ \hline
\multirow{4}{*}{fr} & P@1  & 54.26            & 79.17             & 48.17/52.83/53.47 & 54.65/60.38/61.81 & 45.40/49.59/49.90   & 45.58/50.84/50.84 \\
                    & P@5  & 62.78            & 88.12             & 61.49/62.92/63.56 & 68.26/71.12/71.12 & 58.59/59.41/60.33   & 60.14/61.10/62.29 \\
                    & P@10 & 66.43            & 90.69             & 64.64/66.07/66.50 & 71.36/72.55/73.03 & 61.76/63.29/63.70   & 63.96/66.11/66.59 \\
                    & P@30 & 71.08            & 93.84             & 69.29/71.15/71.73 & 75.66/77.33/77.80 & 66.56/68.51/69.12   & 69.21/71.36/71.60 \\ \hline
\multirow{4}{*}{ro} & P@1  & 45.96            & 80.6              & 42.30/45.88/47.46 & 48.45/53.22/54.89 & 39.67/42.74/44.27   & 41.53/45.35/46.30 \\
                    & P@5  & 56.76            & 89.62             & 54.90/58.70/59.99 & 61.58/66.35/67.78 & 52.04/55.42/56.65   & 55.13/56.56/58.23 \\
                    & P@10 & 60.92            & 91.48             & 60.06/62.56/63.92 & 65.39/70.41/71.12 & 57.77/59.20/60.84   & 59.19/60.14/61.58 \\
                    & P@30 & 67.07            & 94.63             & 66.79/68.22/68.86 & 73.27/74.46/73.99 & 64.01/65.54/66.67   & 65.39/66.59/67.78 \\ \hline
\multirow{4}{*}{eu} & P@1  & 34.86            & 77.81             & 25.63/32.21/33.43 & 29.59/38.66/40.10 & 23.93/29.45/30.57   & 24.34/29.83/30.79 \\
                    & P@5  & 43.52            & 87.97             & 39.58/42.45/44.09 & 46.78/51.07/53.22 & 36.50/38.75/40.18   & 37.23/39.62/40.33 \\
                    & P@10 & 46.67            & 90.34             & 42.30/45.31/47.32 & 50.36/53.70/56.09 & 38.85/41.72/43.56   & 39.38/42.24/43.68 \\
                    & P@30 & 52.11            & 92.63             & 48.39/50.75/53.19 & 56.32/60.14/62.77 & 44.99/46.73/49.08   & 44.39/47.02/49.40 \\ \hline
\multirow{4}{*}{fi} & P@1  & 39.08            & 76.95             & 30.28/34.79/36.72 & 39.62/42.96/46.06 & 26.28/31.29/32.72   & 27.45/30.79/32.22 \\
                    & P@5  & 50.47            & 86.97             & 44.67/47.53/49.03 & 53.70/57.28/59.67 & 40.80/43.35/44.48   & 42.72/44.63/46.06 \\
                    & P@10 & 53.54            & 89.62             & 48.46/51.40/52.54 & 57.28/62.53/63.48 & 44.68/46.63/47.85   & 46.54/48.69/50.12 \\
                    & P@30 & 58.77            & 92.56             & 54.19/56.62/58.34 & 64.92/67.30/68.74 & 49.59/52.04/53.89   & 52.51/54.18/55.61 \\ \hline
\multirow{4}{*}{ja} & P@1  & 41.95            & 79.46             & 24.19/32.93/34.22 & 23.63/32.46/33.65 & 24.44/33.13/34.46   & 25.54/35.80/36.52 \\
                    & P@5  & 57.12            & 89.55             & 47.24/52.61/53.61 & 48.69/54.89/57.28 & 46.63/51.64/52.04   & 52.03/56.09/55.85 \\
                    & P@10 & 61.92            & 92.13             & 54.19/58.48/58.84 & 57.28/60.86/60.86 & 52.86/57.46/57.98   & 58.00/61.10/62.05 \\
                    & P@30 & 67.86            & 95.13             & 61.78/65.00/66.00 & 62.29/66.35/67.30 & 61.55/64.42/65.44   & 65.16/66.35/67.30 \\ \hline
\multirow{4}{*}{th} & P@1  & 38.01            & 80.31             & 23.55/28.13/31.78 & 23.15/29.59/33.65 & 23.72/27.51/30.98   & 25.54/27.92/33.65 \\
                    & P@5  & 56.84            & 90.12             & 49.82/54.26/56.19 & 51.55/58.00/59.43 & 49.08/52.66/54.81   & 51.07/55.85/58.00 \\
                    & P@10 & 62.78            & 93.13             & 57.05/60.99/62.71 & 59.67/63.72/65.39 & 55.93/59.82/61.55   & 57.28/63.25/65.63 \\
                    & P@30 & 70.44            & 96.13             & 67.57/70.22/71.08 & 69.93/72.55/72.55 & 66.56/69.22/70.45   & 68.97/71.36/72.79 \\ \hline
\end{tabular}%
}
\caption{Full results for Aya101 (13B) with average embedding}
\label{tab:my-table}
\end{table*}

\begin{table*}[]
\centering
\resizebox{0.8\textwidth}{!}{%
\begin{tabular}{|c|l|c|c|c|c|c|c|}
\hline
                    & \multicolumn{7}{c|}{\begin{tabular}[c]{@{}c@{}}EN\\ (P@K: 1000/2000/3000) \end{tabular}}                                   \\ \hline
Language            & P@K  & Before-Alignment & Ceiling Performance & Eval - Both       & Eval - Abstract   & Eval-Physical (all) & Eval - Physical (downsample)   \\ \hline
\multirow{4}{*}{fr} & P@1  & 40.52            & 56.12             & 45.96/47.10/47.67 & 56.56/57.52/58.47 & 41.41/42.64/43.05   & 42.24/43.91/44.63 \\
                    & P@5  & 49.82            & 63.92             & 55.26/56.84/57.34 & 66.35/67.78/67.78 & 50.51/52.15/52.86   & 51.07/53.70/54.18 \\
                    & P@10 & 53.76            & 66.14             & 58.48/59.34/59.70 & 68.74/69.45/69.69 & 54.09/55.01/55.42   & 55.61/56.56/57.04 \\
                    & P@30 & 59.91            & 70.15             & 64.14/64.64/64.71 & 73.75/74.22/74.94 & 60.02/60.53/60.33   & 63.25/63.72/63.72 \\ \hline
\multirow{4}{*}{ro} & P@1  & 20.19            & 50.32             & 37.01/39.01/40.01 & 49.64/51.79/52.27 & 31.60/33.54/34.76   & 32.70/35.08/36.75 \\
                    & P@5  & 29.99            & 59.2              & 47.89/49.75/50.54 & 60.38/63.01/63.25 & 42.54/44.07/45.09   & 44.39/46.30/46.30 \\
                    & P@10 & 35.22            & 61.78             & 52.11/53.40/54.26 & 64.44/66.35/66.83 & 46.83/47.85/48.88   & 48.45/49.40/50.36 \\
                    & P@30 & 43.45            & 66.71             & 57.98/59.41/59.77 & 70.88/73.27/73.51 & 52.45/53.48/53.89   & 53.46/54.65/54.65 \\ \hline
\multirow{4}{*}{eu} & P@1  & 11.81            & 30.99             & 17.18/19.18/20.62 & 23.63/26.25/27.68 & 14.42/16.16/17.59   & 15.75/17.42/18.38 \\
                    & P@5  & 19.83            & 38.8              & 26.27/27.77/28.92 & 35.32/37.71/38.42 & 22.39/23.52/24.85   & 23.87/25.06/26.01 \\
                    & P@10 & 23.34            & 42.38             & 30.28/31.85/32.50 & 41.29/42.72/43.20 & 25.56/27.20/27.91   & 26.97/28.88/29.83 \\
                    & P@30 & 29.78            & 47.17             & 36.36/37.87/38.37 & 46.06/48.21/49.64 & 32.21/33.44/33.54   & 32.94/33.65/34.13 \\ \hline
\multirow{4}{*}{fi} & P@1  & 26.34            & 43.66             & 29.35/31.78/32.78 & 37.71/39.86/40.81 & 25.77/28.32/29.35   & 26.97/29.12/30.79 \\
                    & P@5  & 37.29            & 52.68             & 41.02/42.16/42.73 & 51.79/53.22/53.94 & 36.40/37.42/37.93   & 35.80/37.71/37.23 \\
                    & P@10 & 41.88            & 56.91             & 45.03/46.39/46.96 & 56.56/58.00/58.95 & 40.08/41.41/41.82   & 40.10/40.33/40.10 \\
                    & P@30 & 49.68            & 62.99             & 52.61/53.40/54.19 & 62.77/63.25/64.44 & 48.26/49.18/49.80   & 47.02/48.21/48.21 \\ \hline
\multirow{4}{*}{ja} & P@1  & 19.54            & 42.81             & 25.13/27.63/27.70 & 33.17/36.99/36.28 & 21.68/23.62/24.03   & 22.91/26.01/26.73 \\
                    & P@5  & 31.85            & 54.83             & 38.80/40.44/41.66 & 48.93/49.88/50.60 & 34.46/36.40/37.83   & 36.99/38.42/40.57 \\
                    & P@10 & 37.22            & 58.91             & 43.88/45.88/46.67 & 53.70/55.37/55.37 & 39.67/41.82/42.94   & 41.53/43.68/45.11 \\
                    & P@30 & 46.17            & 65.14             & 52.83/54.33/54.97 & 61.81/63.96/65.39 & 48.98/50.20/50.51   & 49.88/51.31/51.79 \\ \hline
\multirow{4}{*}{th} & P@1  & 17.82            & 44.52             & 24.70/26.70/27.77 & 27.21/30.31/32.70 & 23.62/25.15/25.66   & 26.01/27.21/29.12 \\
                    & P@5  & 29.99            & 57.19             & 40.59/42.30/42.81 & 48.93/50.60/51.79 & 37.01/38.75/38.96   & 37.95/39.38/39.86 \\
                    & P@10 & 36.65            & 61.42             & 46.53/48.03/47.89 & 54.18/56.09/56.80 & 43.25/44.58/44.07   & 44.87/46.54/46.54 \\
                    & P@30 & 45.81            & 69.58             & 55.69/56.69/57.48 & 63.96/64.20/65.39 & 52.15/53.48/54.09   & 54.65/56.32/57.04 \\ \hline
\end{tabular}%
}
\caption{Full results for Aya101 (13B) with prompt-based embedding}
\label{tab:my-table}
\end{table*}

\begin{table*}[]
\centering
\resizebox{0.8\textwidth}{!}{%
\begin{tabular}{|c|l|c|c|c|c|c|c|}
\hline
                    & \multicolumn{7}{c|}{\begin{tabular}[c]{@{}c@{}}EN\\ (P@K: 1000/2000/3000) \end{tabular}}                                   \\ \hline
Language            & P@K  & Before-Alignment & Ceiling Performance & Eval - Both       & Eval - Abstract   & Eval-Physical (all) & Eval - Physical (downsample)   \\ \hline
\multirow{4}{*}{fr} & P@1  & 24.27            & 34.29             & 22.62/25.77/27.20 & 24.34/29.12/30.55 & 21.88/24.34/25.77   & 21.96/23.87/26.01 \\
                    & P@5  & 30.42            & 40.8              & 30.06/33.21/34.36 & 34.61/38.42/39.86 & 28.12/30.98/32.00   & 29.12/31.50/32.70 \\
                    & P@10 & 33.57            & 43.24             & 32.93/35.22/36.51 & 38.19/40.81/41.77 & 30.67/32.82/34.25   & 32.22/33.41/34.37 \\
                    & P@30 & 38.73            & 47.82             & 38.01/40.23/41.30 & 42.96/46.06/47.49 & 35.89/37.73/38.65   & 36.99/39.14/40.57 \\ \hline
\multirow{4}{*}{ro} & P@1  & 14.17            & 18.18             & 8.02/10.31/11.24  & 6.44/9.07/9.79    & 8.69/10.84/11.86    & 8.59/10.98/12.65  \\
                    & P@5  & 16.75            & 22.98             & 13.82/15.60/16.75 & 12.17/14.56/16.23 & 14.52/16.05/16.87   & 15.75/17.66/18.38 \\
                    & P@10 & 18.25            & 24.48             & 16.25/17.47/18.61 & 15.51/17.42/17.90 & 16.56/17.48/18.92   & 17.42/19.09/20.53 \\
                    & P@30 & 20.97            & 28.2              & 20.04/20.83/21.90 & 19.81/21.00/21.72 & 20.14/20.76/21.98   & 21.48/22.43/23.87 \\ \hline
\multirow{4}{*}{eu} & P@1  & 9.74             & 15.32             & 5.65/8.45/9.16    & 6.68/10.26/11.22  & 5.21/7.67/8.28      & 4.53/6.92/7.40    \\
                    & P@5  & 12.67            & 20.47             & 10.88/12.88/13.82 & 12.41/14.56/15.27 & 10.22/12.17/13.19   & 10.02/11.22/11.69 \\
                    & P@10 & 14.1             & 21.9              & 12.96/15.25/15.82 & 15.27/17.90/17.90 & 11.96/14.11/14.93   & 11.93/12.65/13.13 \\
                    & P@30 & 17.32            & 25.2              & 16.39/17.97/19.04 & 19.33/21.00/21.72 & 15.13/16.67/17.89   & 14.56/15.51/15.99 \\ \hline
\multirow{4}{*}{fi} & P@1  & 5.15             & 7.09              & 1.36/2.00/2.51    & 1.91/2.39/3.10    & 1.12/1.84/2.25      & 0.48/1.19/1.67    \\
                    & P@5  & 5.8              & 10.74             & 2.51/4.01/4.44    & 2.63/3.82/5.01    & 2.45/4.09/4.19      & 2.63/4.06/3.82    \\
                    & P@10 & 6.73             & 12.46             & 3.87/5.51/5.94    & 4.53/6.21/7.16    & 3.58/5.21/5.42      & 4.06/5.01/5.49    \\
                    & P@30 & 8.45             & 15.75             & 6.51/7.95/8.45    & 6.92/8.11/9.07    & 6.34/7.77/8.18      & 6.68/7.40/7.88    \\ \hline
\multirow{4}{*}{ja} & P@1  & 0.07             & 4.72              & 0.36/0.57/0.86    & 0.24/0.48/0.72    & 0.41/0.61/0.92      & 0.00/0.48/0.48    \\
                    & P@5  & 0.14             & 8.09              & 1.29/2.51/2.58    & 1.67/3.58/3.34    & 1.12/2.04/2.25      & 0.72/1.19/1.43    \\
                    & P@10 & 0.64             & 10.52             & 2.43/3.72/3.65    & 3.58/4.53/4.77    & 1.94/3.37/3.17      & 1.19/2.15/1.91    \\
                    & P@30 & 1.86             & 16.46             & 5.87/6.51/6.80    & 7.88/8.83/8.59    & 5.01/5.52/6.03      & 3.58/4.30/4.77    \\ \hline
\multirow{4}{*}{th} & P@1  & 0                & 3.01              & 0.14/0.21/0.21    & 0.00/0.48/0.48    & 0.20/0.10/0.10      & 0.00/0.24/0.00    \\
                    & P@5  & 0.14             & 5.94              & 0.29/0.72/0.79    & 0.24/1.43/1.43    & 0.31/0.41/0.51      & 0.24/0.24/0.24    \\
                    & P@10 & 0.21             & 7.66              & 0.50/1.00/1.36    & 0.48/1.91/2.63    & 0.51/0.61/0.82      & 0.24/0.24/0.48    \\
                    & P@30 & 1.5              & 12.53             & 2.22/2.79/2.65    & 4.30/5.25/4.77    & 1.33/1.74/1.74      & 0.72/1.67/1.43 \\ \hline
\end{tabular}%
}
\caption{Full results for BLOOMZ-1B7 with last-token embedding}
\label{tab:my-table}
\end{table*}

\begin{table*}[]
\centering
\resizebox{0.8\textwidth}{!}{%
\begin{tabular}{|c|l|c|c|c|c|c|c|}
\hline
                    & \multicolumn{7}{c|}{\begin{tabular}[c]{@{}c@{}}EN\\ (P@K: 1000/2000/3000) \end{tabular}}                                   \\ \hline
Language            & P@K  & Before-Alignment & Ceiling Performance & Eval - Both       & Eval - Abstract   & Eval-Physical (all) & Eval - Physical (downsample)   \\ \hline
\multirow{4}{*}{fr} & P@1  & 24.55            & 48.17             & 42.30/44.38/44.81 & 46.06/49.64/50.12 & 40.70/42.13/42.54   & 40.81/42.48/42.72 \\
                    & P@5  & 36.72            & 58.12             & 51.25/52.76/53.61 & 58.71/60.62/62.29 & 48.06/49.39/49.90   & 50.12/51.07/51.07 \\
                    & P@10 & 43.02            & 61.27             & 55.12/56.34/57.55 & 63.25/64.44/65.16 & 51.64/52.86/54.29   & 53.22/54.65/56.09 \\
                    & P@30 & 52.11            & 65.78             & 60.70/61.99/62.49 & 67.30/68.26/68.26 & 57.87/59.30/60.02   & 59.19/60.62/61.81 \\ \hline
\multirow{4}{*}{ro} & P@1  & 13.89            & 28.2              & 18.68/20.33/21.40 & 19.57/21.24/22.43 & 18.30/19.94/20.96   & 19.09/20.53/21.48 \\
                    & P@5  & 20.62            & 35.93             & 26.34/28.27/29.35 & 28.40/30.79/31.50 & 25.46/27.20/28.43   & 25.54/27.92/28.40 \\
                    & P@10 & 23.48            & 38.51             & 29.35/30.99/32.14 & 31.74/32.22/33.17 & 28.32/30.47/31.70   & 29.36/31.50/32.46 \\
                    & P@30 & 27.7             & 42.66             & 33.57/36.08/37.01 & 35.56/36.75/37.23 & 32.72/35.79/36.91   & 33.41/36.28/36.99 \\ \hline
\multirow{4}{*}{eu} & P@1  & 3.01             & 12.81             & 5.73/7.16/8.16    & 6.44/7.16/9.07    & 5.42/7.16/7.77      & 4.06/5.25/5.73    \\
                    & P@5  & 6.16             & 17.9              & 11.38/13.03/13.67 & 12.89/14.56/15.27 & 10.74/12.37/12.99   & 9.07/11.69/11.93  \\
                    & P@10 & 7.59             & 19.83             & 13.39/15.46/15.96 & 14.80/16.95/18.14 & 12.78/14.83/15.03   & 10.98/14.80/14.80 \\
                    & P@30 & 10.67            & 25.27             & 17.54/18.90/19.47 & 19.09/21.24/22.67 & 16.87/17.89/18.10   & 15.51/16.23/16.71 \\ \hline
\multirow{4}{*}{fi} & P@1  & 6.87             & 11.31             & 3.94/5.80/6.87    & 4.53/6.68/7.64    & 3.68/5.42/6.54      & 5.01/5.97/6.68    \\
                    & P@5  & 10.31            & 16.03             & 7.59/9.59/10.95   & 8.59/10.50/11.69  & 7.16/9.20/10.63     & 9.07/10.02/10.74  \\
                    & P@10 & 11.81            & 18.04             & 9.52/12.17/12.81  & 10.50/12.65/12.41 & 9.10/11.96/12.99    & 10.98/11.93/12.65 \\
                    & P@30 & 13.24            & 22.48             & 13.17/15.39/15.96 & 13.84/15.27/15.27 & 12.88/15.44/16.26   & 14.80/15.51/16.47 \\ \hline
\multirow{4}{*}{ja} & P@1  & 0.07             & 11.1              & 3.87/4.80/5.73    & 7.16/8.59/9.79    & 2.45/3.17/3.99      & 2.15/3.58/4.53    \\
                    & P@5  & 0.5              & 19.26             & 8.45/10.31/11.60  & 12.89/15.75/15.99 & 6.54/7.98/9.71      & 6.68/8.59/10.50   \\
                    & P@10 & 0.86             & 22.69             & 11.17/13.82/14.60 & 15.27/19.33/19.57 & 9.41/11.45/12.47    & 9.79/12.17/12.65  \\
                    & P@30 & 2.29             & 30.49             & 18.32/19.90/20.83 & 22.67/26.25/26.25 & 16.46/17.18/18.51   & 16.71/16.47/17.18 \\ \hline
\multirow{4}{*}{th} & P@1  & 0                & 2.65              & 0.21/0.21/0.50    & 0.48/0.48/0.95    & 0.10/0.10/0.31      & 0.24/0.24/0.48    \\
                    & P@5  & 0                & 5.37              & 0.93/1.43/1.65    & 0.95/2.15/2.86    & 0.92/1.12/1.12      & 1.19/1.91/1.19    \\
                    & P@10 & 0.14             & 7.8               & 1.65/2.79/2.93    & 2.39/4.06/4.06    & 1.33/2.25/2.45      & 1.67/2.86/3.10    \\
                    & P@30 & 1.43             & 12.31             & 4.58/5.08/5.87    & 6.44/6.92/7.64    & 3.78/4.29/5.11      & 3.82/5.25/5.73  \\ \hline
\end{tabular}%
}
\caption{Full results for BLOOMZ-1B7 with prompt-based embedding}
\label{tab:my-table}
\end{table*}

\begin{table*}[]
\centering
\resizebox{0.8\textwidth}{!}{%
\begin{tabular}{|c|l|c|c|c|c|c|c|}
\hline
                    & \multicolumn{7}{c|}{\begin{tabular}[c]{@{}c@{}}EN\\ (P@K: 1000/2000/3000) \end{tabular}}                                   \\ \hline
Language            & P@K  & Before-Alignment & Ceiling Performance & Eval - Both       & Eval - Abstract   & Eval-Physical (all) & Eval - Physical (downsample)   \\ \hline
\multirow{4}{*}{fr} & P@1  & 22.91            & 36.94             & 21.69/25.84/27.27 & 24.34/30.31/31.26 & 20.55/23.93/25.56   & 17.42/21.24/22.91 \\
                    & P@5  & 26.63            & 44.52             & 30.92/34.29/35.72 & 33.89/39.38/40.57 & 29.65/32.11/33.64   & 26.49/28.64/30.55 \\
                    & P@10 & 29.56            & 47.67             & 34.50/37.80/39.37 & 39.14/43.44/45.35 & 32.52/35.38/36.81   & 29.83/33.41/34.84 \\
                    & P@30 & 35.43            & 53.19             & 40.09/42.88/44.09 & 46.78/51.07/51.31 & 37.22/39.37/41.00   & 35.80/38.66/41.29 \\ \hline
\multirow{4}{*}{ro} & P@1  & 14.17            & 16.68             & 4.65/6.37/7.30    & 4.30/6.44/7.64    & 4.81/6.34/7.16      & 6.44/7.88/8.59    \\
                    & P@5  & 16.61            & 21.98             & 9.74/11.67/12.60  & 10.74/12.17/12.65 & 9.30/11.45/12.58    & 11.22/12.89/14.08 \\
                    & P@10 & 17.32            & 24.27             & 12.03/13.82/14.96 & 12.89/15.04/14.80 & 11.66/13.29/15.03   & 12.89/14.56/16.23 \\
                    & P@30 & 19.33            & 28.85             & 17.04/18.47/19.69 & 17.66/18.85/20.29 & 16.77/18.30/19.43   & 17.90/20.29/21.00 \\ \hline
\multirow{4}{*}{eu} & P@1  & 11.38            & 16.68             & 2.43/4.80/5.73    & 3.82/6.92/7.16    & 1.84/3.89/5.11      & 1.91/4.53/6.21    \\
                    & P@5  & 14.46            & 22.76             & 7.95/10.45/12.60  & 10.50/12.65/14.56 & 6.85/9.51/11.76     & 7.40/10.74/12.65  \\
                    & P@10 & 16.46            & 24.62             & 10.16/13.53/14.39 & 13.60/15.51/17.18 & 8.79/12.68/13.19    & 9.55/12.89/14.08  \\
                    & P@30 & 18.9             & 29.85             & 15.60/17.11/18.75 & 19.33/19.33/21.48 & 14.01/16.16/17.59   & 14.08/16.47/18.14 \\ \hline
\multirow{4}{*}{fi} & P@1  & 4.94             & 8.09              & 0.72/1.07/1.65    & 1.91/2.15/2.15    & 0.20/0.61/1.43      & 0.24/0.48/1.19    \\
                    & P@5  & 6.37             & 12.6              & 2.65/3.51/4.08    & 4.30/5.25/5.49    & 1.94/2.76/3.48      & 1.67/2.39/3.10    \\
                    & P@10 & 7.23             & 14.46             & 3.87/4.65/5.58    & 5.97/6.92/7.64    & 2.97/3.68/4.70      & 2.63/3.58/4.30    \\
                    & P@30 & 8.66             & 18.75             & 6.87/6.87/8.30    & 9.79/10.02/11.69  & 5.62/5.52/6.85      & 5.73/5.97/7.16    \\ \hline
\multirow{4}{*}{ja} & P@1  & 0.21             & 5.87              & 0.43/0.79/0.72    & 0.72/0.72/0.48    & 0.31/0.82/0.82      & 0.24/0.72/0.48    \\
                    & P@5  & 0.93             & 10.31             & 1.72/2.00/2.15    & 2.39/2.63/2.63    & 1.43/1.74/1.94      & 0.72/0.95/1.43    \\
                    & P@10 & 1.36             & 13.53             & 2.79/3.44/3.72    & 3.58/5.25/5.97    & 2.45/2.66/2.76      & 2.39/2.86/2.86    \\
                    & P@30 & 2.15             & 19.11             & 7.09/7.44/8.38    & 8.11/10.98/12.41  & 6.65/5.93/6.65      & 6.68/6.21/7.64    \\ \hline
\multirow{4}{*}{th} & P@1  & 0                & 5.01              & 0.14/0.21/0.29    & 0.24/0.24/0.48    & 0.10/0.20/0.20      & 0.00/0.24/0.24    \\
                    & P@5  & 0.29             & 8.16              & 0.72/1.07/0.93    & 0.72/1.43/1.67    & 0.72/0.92/0.61      & 0.48/1.19/0.95    \\
                    & P@10 & 0.29             & 10.52             & 1.15/1.57/1.72    & 1.19/2.63/2.86    & 1.12/1.12/1.23      & 0.95/1.43/1.43    \\
                    & P@30 & 0.72             & 15.39             & 2.58/3.44/3.79    & 3.34/6.21/6.21    & 2.25/2.25/2.76      & 1.67/2.15/2.39 \\ \hline
\end{tabular}%
}
\caption{Full results for BLOOMZ-3B with last-token embedding}
\label{tab:my-table}
\end{table*}

\begin{table*}[]
\centering
\resizebox{0.8\textwidth}{!}{%
\begin{tabular}{|c|l|c|c|c|c|c|c|}
\hline
                    & \multicolumn{7}{c|}{\begin{tabular}[c]{@{}c@{}}EN\\ (P@K: 1000/2000/3000) \end{tabular}}                                   \\ \hline
Language            & P@K  & Before-Alignment & Ceiling Performance & Eval - Both       & Eval - Abstract   & Eval-Physical (all) & Eval - Physical (downsample)   \\ \hline
\multirow{4}{*}{fr} & P@1  & 19.33            & 58.48             & 52.18/53.33/53.69 & 59.90/60.38/60.38 & 48.88/50.31/50.82   & 48.93/50.36/50.60 \\
                    & P@5  & 34.93            & 69.29             & 63.28/64.07/64.71 & 71.36/72.55/73.03 & 59.82/60.43/61.15   & 61.34/62.29/62.77 \\
                    & P@10 & 41.95            & 72.15             & 66.28/67.43/67.86 & 73.75/75.42/76.13 & 63.09/64.01/64.31   & 65.87/66.11/66.59 \\
                    & P@30 & 55.48            & 77.38             & 71.22/72.08/73.09 & 78.76/79.24/79.95 & 68.00/69.02/70.14   & 70.41/71.36/72.79 \\ \hline
\multirow{4}{*}{ro} & P@1  & 2.43             & 33.43             & 21.33/23.91/26.27 & 24.58/26.97/28.64 & 19.94/22.60/25.26   & 21.72/22.91/25.06 \\
                    & P@5  & 6.59             & 41.95             & 30.99/33.07/35.36 & 34.84/36.99/37.47 & 29.35/31.39/34.46   & 31.26/32.46/35.56 \\
                    & P@10 & 9.66             & 45.24             & 34.22/36.29/37.65 & 37.71/38.90/39.86 & 32.72/35.17/36.71   & 35.32/37.71/37.23 \\
                    & P@30 & 15.39            & 49.75             & 40.73/43.02/43.52 & 44.15/44.15/46.06 & 39.26/42.54/42.43   & 40.81/43.91/43.68 \\ \hline
\multirow{4}{*}{eu} & P@1  & 1.07             & 18.75             & 8.95/11.17/11.24  & 10.02/14.32/14.32 & 8.49/9.82/9.92      & 8.11/10.02/10.26  \\
                    & P@5  & 3.79             & 26.34             & 15.25/18.25/19.61 & 18.62/23.63/25.06 & 13.80/15.95/17.28   & 14.56/16.47/17.66 \\
                    & P@10 & 5.08             & 29.06             & 18.18/21.62/22.55 & 21.72/26.49/27.45 & 16.67/19.53/20.45   & 18.14/18.85/20.76 \\
                    & P@30 & 9.23             & 34.86             & 23.62/26.41/27.27 & 28.16/30.31/31.26 & 21.68/24.74/25.56   & 23.63/24.82/25.78 \\ \hline
\multirow{4}{*}{fi} & P@1  & 1.22             & 13.67             & 5.58/7.09/8.23    & 7.40/9.31/11.22   & 4.81/6.13/6.95      & 5.25/6.21/7.16    \\
                    & P@5  & 2.51             & 19.18             & 9.81/11.31/12.53  & 11.69/14.08/14.56 & 9.00/10.12/11.66    & 10.02/10.26/11.69 \\
                    & P@10 & 3.36             & 21.69             & 11.38/13.60/14.82 & 12.89/15.99/16.71 & 10.74/12.58/14.01   & 10.98/12.89/14.56 \\
                    & P@30 & 6.37             & 25.34             & 13.96/16.68/18.32 & 15.99/18.38/20.05 & 13.09/15.95/17.59   & 13.13/16.95/19.57 \\ \hline
\multirow{4}{*}{ja} & P@1  & 0.79             & 13.17             & 3.29/4.65/5.73    & 6.21/9.07/11.22   & 2.04/2.76/3.37      & 2.39/2.86/3.10    \\
                    & P@5  & 2.86             & 22.98             & 11.31/12.38/14.10 & 17.90/19.81/22.43 & 8.49/9.20/10.53     & 9.07/9.07/9.79    \\
                    & P@10 & 4.44             & 26.7              & 14.46/16.03/17.68 & 21.24/23.87/27.68 & 11.55/12.68/13.39   & 11.22/12.41/12.89 \\
                    & P@30 & 9.09             & 34.22             & 20.76/23.19/24.19 & 28.88/34.37/35.08 & 17.28/18.40/19.53   & 16.71/16.71/19.33 \\ \hline
\multirow{4}{*}{th} & P@1  & 0.14             & 2.43              & 0.29/0.07/0.29    & 0.24/0.00/0.72    & 0.31/0.10/0.10      & 0.24/0.24/0.24    \\
                    & P@5  & 0.43             & 6.37              & 1.29/1.22/0.86    & 1.67/1.43/0.95    & 1.12/1.12/0.82      & 0.95/0.95/0.48    \\
                    & P@10 & 0.64             & 9.74              & 1.57/1.72/1.65    & 1.91/2.63/2.39    & 1.43/1.33/1.33      & 1.43/1.19/0.95    \\
                    & P@30 & 1.86             & 15.32             & 2.93/4.44/5.08    & 3.34/7.16/7.40    & 2.76/3.27/4.09      & 3.34/3.34/4.06 \\ \hline
\end{tabular}%
}
\caption{Full results for BLOOMZ-3B with prompt-based embedding}
\label{tab:my-table}
\end{table*}

\begin{table*}[]
\centering
\resizebox{0.8\textwidth}{!}{%
\begin{tabular}{|c|l|c|c|c|c|c|c|}
\hline
                    & \multicolumn{7}{c|}{\begin{tabular}[c]{@{}c@{}}EN\\ (P@K: 1000/2000/3000) \end{tabular}}                                   \\ \hline
Language            & P@K  & Before-Alignment & Ceiling Performance & Eval - Both       & Eval - Abstract   & Eval-Physical (all) & Eval - Physical (downsample)   \\ \hline
\multirow{4}{*}{fr} & P@1  & 17.61            & 34.5              & 16.82/19.83/21.19 & 21.96/25.30/25.54 & 14.62/17.48/19.33   & 13.60/16.71/17.42 \\
                    & P@5  & 20.33            & 43.24             & 25.98/28.35/30.21 & 31.03/34.61/36.28 & 23.82/25.66/27.61   & 23.39/25.06/26.73 \\
                    & P@10 & 21.26            & 47.03             & 29.92/32.21/34.86 & 35.56/38.19/40.33 & 27.51/29.65/32.52   & 26.97/28.88/32.22 \\
                    & P@30 & 23.77            & 51.9              & 37.51/40.80/41.45 & 43.68/46.54/47.49 & 34.87/38.34/38.85   & 35.08/38.90/39.86 \\ \hline
\multirow{4}{*}{ro} & P@1  & 13.24            & 19.11             & 3.79/5.65/7.09    & 4.06/4.77/5.97    & 3.68/6.03/7.57      & 4.06/7.16/8.59    \\
                    & P@5  & 14.96            & 25.2              & 8.38/12.24/13.53  & 8.35/10.98/11.69  & 8.38/12.78/14.31    & 9.07/14.56/15.75  \\
                    & P@10 & 16.39            & 28.49             & 11.74/14.60/15.53 & 10.98/13.84/13.60 & 12.07/14.93/16.36   & 13.13/15.99/17.42 \\
                    & P@30 & 18.18            & 33.43             & 16.82/19.90/21.05 & 16.47/17.90/19.09 & 16.97/20.76/21.88   & 17.90/21.48/21.96 \\ \hline
\multirow{4}{*}{eu} & P@1  & 7.87             & 17.32             & 3.65/5.30/6.01    & 3.82/5.97/6.92    & 3.58/5.01/5.62      & 4.06/5.25/5.97    \\
                    & P@5  & 9.16             & 24.41             & 7.59/9.88/10.52   & 8.35/11.22/11.69  & 7.26/9.30/10.02     & 7.16/10.02/10.98  \\
                    & P@10 & 10.16            & 26.77             & 10.67/12.88/13.53 & 11.46/14.56/15.04 & 10.33/12.17/12.88   & 10.02/12.65/13.84 \\
                    & P@30 & 12.24            & 31.64             & 16.54/18.68/19.69 & 17.42/20.76/21.00 & 16.16/17.79/19.12   & 14.56/16.47/17.90 \\ \hline
\multirow{4}{*}{fi} & P@1  & 4.22             & 9.66              & 0.64/0.86/1.15    & 0.72/0.72/1.67    & 0.61/0.92/0.92      & 0.48/0.48/0.72    \\
                    & P@5  & 4.8              & 14.03             & 2.15/3.15/3.94    & 2.86/4.06/5.49    & 1.84/2.76/3.27      & 1.19/1.67/2.86    \\
                    & P@10 & 5.37             & 16.03             & 3.72/4.65/5.30    & 5.01/5.49/6.92    & 3.17/4.29/4.60      & 2.39/3.34/4.30    \\
                    & P@30 & 6.73             & 21.05             & 6.80/8.09/9.09    & 7.16/10.26/10.50  & 6.65/7.16/8.49      & 6.44/6.92/9.07    \\ \hline
\multirow{4}{*}{ja} & P@1  & 0.29             & 9.88              & 0.57/0.64/0.93    & 0.72/1.43/1.43    & 0.51/0.31/0.72      & 0.24/0.24/0.24    \\
                    & P@5  & 0.72             & 16.25             & 2.08/2.51/2.86    & 3.58/3.82/4.30    & 1.43/1.94/2.25      & 1.19/2.39/1.91    \\
                    & P@10 & 1.15             & 18.75             & 3.36/4.08/4.94    & 4.30/5.97/6.92    & 2.97/3.27/4.09      & 2.39/2.86/3.58    \\
                    & P@30 & 2                & 24.84             & 6.59/8.59/8.59    & 9.07/13.37/13.37  & 5.52/6.54/6.54      & 4.53/6.68/6.21    \\ \hline
\multirow{4}{*}{th} & P@1  & 0.07             & 11.1              & 0.36/0.29/0.43    & 0.72/0.48/0.72    & 0.20/0.20/0.31      & 0.00/0.24/0.24    \\
                    & P@5  & 0.14             & 16.25             & 0.93/1.43/1.22    & 1.43/2.15/1.91    & 0.72/1.12/0.92      & 0.72/1.19/0.72    \\
                    & P@10 & 0.43             & 18.9              & 1.43/1.86/1.86    & 2.15/3.10/2.86    & 1.12/1.33/1.43      & 1.43/1.43/1.19    \\
                    & P@30 & 1.36             & 24.34             & 3.51/3.08/3.65    & 4.53/5.01/5.25    & 3.07/2.25/2.97      & 3.34/2.15/2.86  \\ \hline
\end{tabular}%
}
\caption{Full results for BLOOMZ-7B1 with last-token embedding}
\label{tab:my-table}
\end{table*}

\begin{table*}[]
\centering
\resizebox{0.8\textwidth}{!}{%
\begin{tabular}{|c|l|c|c|c|c|c|c|}
\hline
                    & \multicolumn{7}{c|}{\begin{tabular}[c]{@{}c@{}}EN\\ (P@K: 1000/2000/3000) \end{tabular}}                                   \\ \hline
Language            & P@K  & Before-Alignment & Ceiling Performance & Eval - Both       & Eval - Abstract   & Eval-Physical (all) & Eval - Physical (downsample)   \\ \hline
\multirow{4}{*}{fr} & P@1  & 2.58             & 61.85             & 51.11/53.83/55.62 & 60.86/63.01/64.92 & 46.93/49.90/51.64   & 47.49/51.55/52.51 \\
                    & P@5  & 7.95             & 71.58             & 64.57/66.21/66.86 & 73.51/74.46/74.70 & 60.74/62.68/63.50   & 63.25/65.16/66.35 \\
                    & P@10 & 14.96            & 74.45             & 67.64/69.08/69.51 & 75.89/76.13/76.37 & 64.11/66.05/66.56   & 66.59/69.69/69.21 \\
                    & P@30 & 29.56            & 79.31             & 73.09/73.44/73.87 & 79.95/79.47/79.71 & 70.14/70.86/71.37   & 72.55/73.27/74.46 \\ \hline
\multirow{4}{*}{ro} & P@1  & 1.65             & 36.58             & 21.76/25.34/28.13 & 28.64/31.50/33.41 & 18.81/22.70/25.87   & 20.05/24.82/27.45 \\
                    & P@5  & 3.79             & 43.88             & 31.50/34.65/36.15 & 38.19/40.81/40.81 & 28.63/32.00/34.15   & 30.31/34.13/35.80 \\
                    & P@10 & 5.3              & 46.53             & 35.50/38.01/39.73 & 41.77/42.96/44.39 & 32.82/35.89/37.73   & 34.37/38.42/39.38 \\
                    & P@30 & 10.45            & 50.32             & 41.16/42.59/43.88 & 45.82/47.26/49.16 & 39.16/40.59/41.62   & 41.29/43.44/43.20 \\ \hline
\multirow{4}{*}{eu} & P@1  & 1.93             & 30.78             & 18.54/20.26/21.40 & 24.11/27.68/27.92 & 16.16/17.08/18.61   & 16.95/17.18/18.62 \\
                    & P@5  & 4.58             & 40.52             & 28.78/30.35/30.78 & 35.32/37.23/38.19 & 25.97/27.40/27.61   & 25.30/26.73/26.97 \\
                    & P@10 & 6.59             & 43.02             & 32.57/33.00/34.36 & 39.38/40.57/42.72 & 29.65/29.75/30.78   & 28.88/28.16/29.12 \\
                    & P@30 & 12.46            & 49.39             & 36.94/38.01/39.01 & 43.91/45.82/47.26 & 33.95/34.66/35.48   & 32.22/33.89/35.32 \\ \hline
\multirow{4}{*}{fi} & P@1  & 0.86             & 15.1              & 5.65/7.30/7.80    & 8.59/10.74/10.74  & 4.40/5.83/6.54      & 4.77/5.73/6.44    \\
                    & P@5  & 2.65             & 20.26             & 10.02/12.67/13.46 & 13.60/15.99/16.23 & 8.49/11.25/12.27    & 9.07/11.69/11.93  \\
                    & P@10 & 3.44             & 22.98             & 12.10/14.60/15.60 & 15.27/17.66/18.14 & 10.74/13.29/14.52   & 11.93/13.37/14.80 \\
                    & P@30 & 6.01             & 27.13             & 15.60/17.90/18.61 & 19.09/21.00/21.24 & 14.11/16.56/17.48   & 15.27/16.23/16.95 \\ \hline
\multirow{4}{*}{ja} & P@1  & 0.5              & 14.32             & 4.01/4.58/5.08    & 7.40/8.83/10.26   & 2.56/2.76/2.86      & 3.34/4.30/4.53    \\
                    & P@5  & 2.43             & 23.34             & 8.09/10.38/11.60  & 14.32/18.62/20.05 & 5.42/6.85/7.98      & 7.88/8.83/10.02   \\
                    & P@10 & 3.44             & 27.92             & 11.45/14.60/15.68 & 18.85/23.39/26.01 & 8.28/10.84/11.25    & 9.79/12.89/13.37  \\
                    & P@30 & 7.8              & 35.79             & 17.04/20.54/22.69 & 25.78/31.50/33.41 & 13.29/15.85/18.10   & 15.27/17.66/19.81 \\ \hline
\multirow{4}{*}{th} & P@1  & 0.14             & 4.44              & 0.50/0.50/0.50    & 0.48/0.95/1.19    & 0.51/0.31/0.20      & 0.48/0.00/0.00    \\
                    & P@5  & 0.79             & 10.67             & 1.65/1.72/2.22    & 1.91/2.39/3.58    & 1.53/1.43/1.64      & 1.91/1.19/0.95    \\
                    & P@10 & 1.22             & 14.03             & 3.01/2.86/3.72    & 4.06/4.30/5.73    & 2.56/2.25/2.86      & 2.39/1.43/2.15    \\
                    & P@30 & 2.65             & 22.12             & 6.23/6.87/7.02    & 7.64/9.79/9.55    & 5.62/5.62/5.93      & 5.97/5.73/5.01 \\ \hline
\end{tabular}%
}
\caption{Full results for BLOOMZ-7B1 with prompt-based embedding}
\label{tab:my-table}
\end{table*}

\begin{table*}[]
\centering
\resizebox{0.8\textwidth}{!}{%
\begin{tabular}{|c|l|c|c|c|c|c|c|}
\hline
                    & \multicolumn{7}{c|}{\begin{tabular}[c]{@{}c@{}}EN\\ (P@K: 1000/2000/3000) \end{tabular}}                                   \\ \hline
Language            & P@K  & Before-Alignment & Ceiling Performance & Eval - Both       & Eval - Abstract   & Eval-Physical (all) & Eval - Physical (downsample)   \\ \hline
\multirow{4}{*}{fr} & P@1  & 35.86            & 49.11             & 33.57/35.22/36.44 & 38.66/40.81/41.29 & 31.39/32.82/34.36   & 31.26/33.89/35.08 \\
                    & P@5  & 45.45            & 60.27             & 43.45/45.03/46.24 & 50.12/52.03/52.03 & 40.59/42.02/43.76   & 42.96/44.39/46.78 \\
                    & P@10 & 48.46            & 63.49             & 46.89/49.11/49.82 & 52.98/55.85/56.32 & 44.27/46.22/47.03   & 47.49/49.88/51.79 \\
                    & P@30 & 53.33            & 69.51             & 52.47/53.83/54.90 & 59.19/59.90/61.34 & 49.59/51.23/52.15   & 53.70/56.09/56.32 \\ \hline
\multirow{4}{*}{ro} & P@1  & 31.35            & 47.24             & 27.63/30.57/32.00 & 31.98/35.32/37.95 & 25.77/28.53/29.45   & 26.97/30.07/30.31 \\
                    & P@5  & 37.72            & 59.41             & 36.22/38.80/39.51 & 42.00/45.35/45.82 & 33.74/35.99/36.81   & 36.52/38.90/40.10 \\
                    & P@10 & 40.87            & 63.56             & 39.01/41.02/42.02 & 45.35/47.02/47.97 & 36.30/38.45/39.47   & 38.42/41.05/42.72 \\
                    & P@30 & 45.17            & 70.15             & 44.60/46.39/47.96 & 50.36/51.55/53.22 & 42.13/44.17/45.71   & 43.91/46.78/48.21 \\ \hline
\multirow{4}{*}{eu} & P@1  & 19.47            & 44.09             & 15.39/18.11/19.90 & 16.47/19.33/22.67 & 14.93/17.59/18.71   & 15.75/18.85/20.29 \\
                    & P@5  & 27.42            & 56.91             & 22.91/25.48/27.27 & 24.58/29.12/31.03 & 22.19/23.93/25.66   & 23.15/25.54/27.68 \\
                    & P@10 & 30.06            & 61.85             & 26.91/29.78/30.49 & 31.03/33.65/36.04 & 25.15/28.12/28.12   & 26.49/29.83/29.59 \\
                    & P@30 & 35               & 69.01             & 33.36/35.29/36.44 & 38.66/41.53/41.77 & 31.08/32.62/34.15   & 32.46/34.13/35.32 \\ \hline
\multirow{4}{*}{fi} & P@1  & 16.96            & 42.45             & 12.53/15.39/17.18 & 12.41/15.51/16.95 & 12.58/15.34/17.28   & 14.08/14.80/17.90 \\
                    & P@5  & 24.12            & 54.04             & 20.90/23.84/25.13 & 21.48/24.58/26.25 & 20.65/23.52/24.64   & 22.91/25.54/26.97 \\
                    & P@10 & 27.34            & 57.62             & 23.84/27.42/28.63 & 25.54/29.83/30.07 & 23.11/26.38/28.02   & 26.25/28.64/31.03 \\
                    & P@30 & 33.07            & 65.21             & 30.64/33.29/35.00 & 33.41/35.80/37.23 & 29.45/32.21/34.05   & 32.46/34.61/37.23 \\ \hline
\multirow{4}{*}{ja} & P@1  & 14.39            & 53.4              & 9.59/13.96/15.60  & 8.11/11.46/12.89  & 10.22/15.03/16.77   & 12.41/19.33/18.85 \\
                    & P@5  & 25.41            & 66.43             & 21.26/27.56/29.42 & 19.33/24.82/26.01 & 22.09/28.73/30.88   & 25.54/32.70/36.28 \\
                    & P@10 & 30.06            & 71.08             & 27.42/33.57/35.79 & 25.30/31.74/33.17 & 28.32/34.36/36.91   & 31.50/39.38/42.24 \\
                    & P@30 & 39.51            & 77.24             & 37.80/43.66/45.17 & 36.99/42.48/44.39 & 38.14/44.17/45.50   & 42.96/50.12/50.84 \\ \hline
\multirow{4}{*}{th} & P@1  & 11.88            & 52.97             & 8.38/10.52/13.10  & 8.83/11.93/13.60  & 8.18/9.92/12.88     & 8.11/10.74/13.84  \\
                    & P@5  & 22.33            & 66.93             & 20.40/24.48/27.70 & 20.53/23.63/25.54 & 20.35/24.85/28.63   & 21.96/27.21/30.55 \\
                    & P@10 & 27.49            & 70.79             & 25.98/31.21/33.50 & 26.73/30.79/33.17 & 25.66/31.39/33.64   & 26.97/33.41/37.47 \\
                    & P@30 & 36.44            & 80.31             & 36.86/41.52/43.24 & 35.56/40.81/43.91 & 37.42/41.82/42.94   & 39.62/43.68/45.58 \\ \hline
\end{tabular}%
}
\caption{Full results for mT0-large (1.2B) with average embedding}
\label{tab:my-table}
\end{table*}

\begin{table*}[]
\centering
\resizebox{0.8\textwidth}{!}{%
\begin{tabular}{|c|l|c|c|c|c|c|c|}
\hline
                    & \multicolumn{7}{c|}{\begin{tabular}[c]{@{}c@{}}EN\\ (P@K: 1000/2000/3000) \end{tabular}}                                   \\ \hline
Language            & P@K  & Before-Alignment & Ceiling Performance & Eval - Both       & Eval - Abstract   & Eval-Physical (all) & Eval - Physical (downsample)   \\ \hline
\multirow{4}{*}{fr} & P@1  & 11.17            & 26.06             & 11.38/14.60/15.60 & 14.32/16.47/17.90 & 9.41/12.37/12.68    & 9.31/10.74/11.46  \\
                    & P@5  & 17.18            & 33.93             & 19.83/22.69/24.41 & 23.15/27.92/30.31 & 17.18/20.25/20.65   & 16.47/19.33/19.09 \\
                    & P@10 & 20.11            & 37.94             & 23.48/26.91/27.99 & 27.45/31.50/36.04 & 21.47/23.52/24.13   & 21.00/23.15/23.15 \\
                    & P@30 & 25.2             & 44.45             & 29.06/33.14/34.43 & 36.04/39.62/42.48 & 25.97/29.86/31.08   & 25.06/28.88/30.79 \\ \hline
\multirow{4}{*}{ro} & P@1  & 6.16             & 14.53             & 3.58/4.15/4.87    & 4.53/7.40/6.44    & 2.86/2.66/3.68      & 3.10/2.86/3.34    \\
                    & P@5  & 11.02            & 24.12             & 7.44/9.59/10.81   & 10.02/14.08/15.51 & 5.83/6.75/7.77      & 6.21/7.40/6.92    \\
                    & P@10 & 13.6             & 28.63             & 10.45/12.88/13.82 & 15.27/18.38/20.53 & 8.28/9.41/10.74     & 9.31/10.50/10.50  \\
                    & P@30 & 18.54            & 36.36             & 16.54/18.40/20.33 & 24.34/27.45/29.12 & 13.50/14.21/15.85   & 15.51/16.47/16.23 \\ \hline
\multirow{4}{*}{eu} & P@1  & 1.29             & 7.09              & 0.64/0.79/1.43    & 0.72/1.19/1.43    & 0.41/0.72/1.12      & 0.24/0.72/0.48    \\
                    & P@5  & 2.93             & 13.96             & 2.15/3.01/3.65    & 3.58/4.53/5.01    & 1.74/2.35/2.86      & 1.19/2.39/3.10    \\
                    & P@10 & 4.15             & 17.11             & 3.08/4.29/5.65    & 4.77/6.21/8.11    & 2.35/3.68/3.89      & 2.15/4.53/4.77    \\
                    & P@30 & 7.52             & 23.84             & 6.66/9.16/10.09   & 9.79/12.17/12.41  & 4.91/7.87/7.98      & 5.25/8.83/9.31    \\ \hline
\multirow{4}{*}{fi} & P@1  & 4.01             & 9.95              & 1.72/2.08/2.29    & 2.63/2.15/2.86    & 0.92/1.33/1.84      & 0.95/1.67/2.15    \\
                    & P@5  & 6.16             & 16.68             & 4.51/4.87/6.94    & 6.92/6.44/10.26   & 2.97/3.68/5.01      & 3.34/4.06/5.97    \\
                    & P@10 & 7.52             & 21.05             & 6.87/7.44/9.09    & 9.31/10.50/13.84  & 5.32/6.13/7.77      & 5.49/6.68/8.59    \\
                    & P@30 & 11.74            & 28.27             & 11.45/12.38/14.03 & 14.80/16.47/18.38 & 9.30/10.63/11.76    & 10.26/10.98/11.93 \\ \hline
\multirow{4}{*}{ja} & P@1  & 0.86             & 7.3               & 0.72/0.93/1.50    & 0.72/1.67/1.67    & 0.61/1.12/0.72      & 0.72/0.95/0.72    \\
                    & P@5  & 1.79             & 15.03             & 2.65/3.29/3.36    & 3.34/3.34/5.01    & 2.04/2.86/2.15      & 3.10/3.58/1.19    \\
                    & P@10 & 3.15             & 18.97             & 4.08/5.08/5.08    & 5.25/6.92/7.16    & 3.99/3.78/3.89      & 5.01/5.01/3.10    \\
                    & P@30 & 5.73             & 26.06             & 7.80/8.95/9.66    & 8.59/11.46/13.13  & 6.75/7.46/8.28      & 7.88/8.83/7.88    \\ \hline
\multirow{4}{*}{th} & P@1  & 0.43             & 7.66              & 0.36/0.64/0.72    & 0.72/0.24/1.19    & 0.31/0.72/0.92      & 0.72/0.72/0.72    \\
                    & P@5  & 1.22             & 15.46             & 2.43/2.58/2.93    & 2.86/3.58/3.58    & 2.45/2.86/2.76      & 2.63/2.86/2.86    \\
                    & P@10 & 1.79             & 19.9              & 4.15/5.30/5.30    & 4.77/6.44/6.92    & 4.19/4.81/4.29      & 3.82/4.53/5.25    \\
                    & P@30 & 3.51             & 26.99             & 8.95/10.31/10.74  & 10.50/10.26/12.17 & 8.28/9.51/9.71      & 7.40/10.26/11.22 \\ \hline
\end{tabular}%
}
\caption{Full results for mT0-large (1.2B) with prompt-based embedding}
\label{tab:my-table}
\end{table*}

\begin{table*}[]
\centering
\resizebox{0.8\textwidth}{!}{%
\begin{tabular}{|c|l|c|c|c|c|c|c|}
\hline
                    & \multicolumn{7}{c|}{\begin{tabular}[c]{@{}c@{}}EN\\ (P@K: 1000/2000/3000) \end{tabular}}                                   \\ \hline
Language            & P@K  & Before-Alignment & Ceiling Performance & Eval - Both       & Eval - Abstract   & Eval-Physical (all) & Eval - Physical (downsample)   \\ \hline
\multirow{4}{*}{fr} & P@1  & 29.35            & 52.76             & 18.83/22.26/24.19 & 18.14/22.91/25.06 & 19.12/21.98/23.82   & 17.42/20.76/23.39 \\
                    & P@5  & 34.29            & 57.77             & 26.49/29.28/30.64 & 26.49/29.36/31.03 & 26.48/29.24/30.47   & 27.45/30.31/31.50 \\
                    & P@10 & 37.01            & 59.2              & 28.42/31.85/32.93 & 28.64/32.22/33.41 & 28.32/31.70/32.72   & 29.59/33.65/34.84 \\
                    & P@30 & 40.94            & 61.92             & 35.36/37.87/38.22 & 37.71/38.90/39.86 & 34.36/37.42/37.53   & 35.80/39.14/39.14 \\ \hline
\multirow{4}{*}{ro} & P@1  & 22.69            & 52.18             & 11.88/15.75/17.90 & 11.22/17.18/18.62 & 12.17/15.13/17.59   & 13.60/15.99/17.90 \\
                    & P@5  & 27.49            & 56.41             & 20.11/23.12/24.91 & 21.24/24.34/26.97 & 19.63/22.60/24.03   & 21.24/22.91/24.34 \\
                    & P@10 & 29.56            & 58.05             & 22.33/25.77/27.77 & 24.11/27.92/29.36 & 21.57/24.85/27.10   & 22.91/25.54/27.45 \\
                    & P@30 & 32.64            & 60.2              & 27.20/29.78/31.64 & 28.88/31.98/32.22 & 26.48/28.83/31.39   & 27.45/28.88/31.03 \\ \hline
\multirow{4}{*}{eu} & P@1  & 15.18            & 54.97             & 5.44/8.88/10.59   & 4.30/9.31/10.50   & 5.93/8.69/10.63     & 5.97/9.31/11.22   \\
                    & P@5  & 19.47            & 59.2              & 12.10/14.75/16.46 & 13.60/16.23/18.62 & 11.45/14.11/15.54   & 12.65/15.27/16.47 \\
                    & P@10 & 21.55            & 60.92             & 14.46/17.54/18.54 & 16.47/18.62/20.53 & 13.60/17.08/17.69   & 14.80/17.42/18.38 \\
                    & P@30 & 24.91            & 62.92             & 19.97/22.91/23.55 & 22.67/26.25/26.49 & 18.81/21.47/22.29   & 20.29/23.63/24.34 \\ \hline
\multirow{4}{*}{fi} & P@1  & 15.32            & 56.34             & 5.08/7.59/9.88    & 5.49/9.07/11.22   & 4.91/6.95/9.30      & 5.25/7.40/9.79    \\
                    & P@5  & 20.26            & 60.06             & 12.38/15.75/16.82 & 11.93/17.90/19.33 & 12.58/14.83/15.75   & 14.56/16.47/17.66 \\
                    & P@10 & 22.41            & 61.13             & 15.25/18.47/20.40 & 16.23/21.24/23.87 & 14.83/17.28/18.92   & 16.47/19.33/19.33 \\
                    & P@30 & 25.55            & 63.64             & 20.11/23.26/24.98 & 21.72/26.49/28.40 & 19.43/21.88/23.52   & 21.24/22.67/23.87 \\ \hline
\multirow{4}{*}{ja} & P@1  & 10.31            & 56.26             & 1.86/4.15/6.51    & 1.91/3.58/5.73    & 1.84/4.40/6.85      & 1.91/3.82/6.44    \\
                    & P@5  & 18.9             & 60.2              & 9.52/14.67/16.68  & 9.31/14.32/16.95  & 9.61/14.83/16.56    & 9.31/14.08/16.71  \\
                    & P@10 & 23.34            & 60.92             & 14.46/19.69/21.12 & 15.51/19.33/20.29 & 14.01/19.84/21.47   & 14.08/20.29/21.96 \\
                    & P@30 & 29.78            & 62.56             & 21.69/26.91/28.42 & 20.29/25.30/25.54 & 22.29/27.61/29.65   & 23.39/29.12/30.55 \\ \hline
\multirow{4}{*}{th} & P@1  & 9.81             & 58.77             & 2.00/3.58/4.72    & 2.15/4.53/5.73    & 1.94/3.17/4.29      & 1.91/4.30/5.97    \\
                    & P@5  & 18.18            & 63.28             & 12.17/14.67/17.68 & 12.17/14.80/16.95 & 12.17/14.62/18.00   & 13.13/15.75/19.09 \\
                    & P@10 & 22.41            & 64.92             & 16.61/20.69/22.62 & 15.27/19.81/21.24 & 17.18/21.06/23.21   & 19.09/21.72/24.34 \\
                    & P@30 & 30.92            & 67.07             & 26.77/29.85/31.50 & 24.58/28.88/28.64 & 27.71/30.27/32.72   & 27.92/30.55/33.17 \\ \hline
\end{tabular}%
}
\caption{Full results for mT0-xl (3.7B) with average embedding}
\label{tab:my-table}
\end{table*}

\begin{table*}[]
\centering
\resizebox{0.8\textwidth}{!}{%
\begin{tabular}{|c|l|c|c|c|c|c|c|}
\hline
                    & \multicolumn{7}{c|}{\begin{tabular}[c]{@{}c@{}}EN\\ (P@K: 1000/2000/3000) \end{tabular}}                                   \\ \hline
Language            & P@K  & Before-Alignment & Ceiling Performance & Eval - Both       & Eval - Abstract   & Eval-Physical (all) & Eval - Physical (downsample)   \\ \hline
\multirow{4}{*}{fr} & P@1  & 34.93            & 49.25             & 40.37/41.30/42.02 & 50.60/51.31/52.03 & 35.99/37.01/37.73   & 32.70/35.32/35.32 \\
                    & P@5  & 46.17            & 59.56             & 52.18/52.90/53.90 & 63.48/63.25/64.92 & 47.34/48.57/49.18   & 46.78/48.21/48.45 \\
                    & P@10 & 50.89            & 63.56             & 56.69/57.55/58.12 & 66.59/67.30/67.54 & 52.45/53.37/54.09   & 52.03/52.98/53.94 \\
                    & P@30 & 58.05            & 69.79             & 63.64/64.14/64.42 & 72.32/73.99/73.51 & 59.82/59.92/60.53   & 60.14/60.38/60.86 \\ \hline
\multirow{4}{*}{ro} & P@1  & 29.85            & 41.8              & 30.64/33.21/34.29 & 37.47/41.77/42.24 & 27.71/29.65/30.88   & 29.59/32.22/33.65 \\
                    & P@5  & 40.01            & 51.83             & 42.02/43.88/45.03 & 53.94/55.37/57.04 & 36.91/38.96/39.88   & 40.33/42.00/43.20 \\
                    & P@10 & 44.31            & 56.41             & 46.89/48.53/48.60 & 59.19/60.14/60.14 & 41.62/43.56/43.66   & 45.58/46.78/46.78 \\
                    & P@30 & 52.11            & 63.56             & 54.19/54.83/55.12 & 63.72/64.44/65.16 & 50.10/50.61/50.82   & 52.03/52.74/53.46 \\ \hline
\multirow{4}{*}{eu} & P@1  & 12.17            & 26.41             & 15.03/16.46/17.39 & 18.14/19.57/21.24 & 13.70/15.13/15.75   & 14.80/15.75/15.75 \\
                    & P@5  & 19.26            & 36.22             & 23.84/25.41/25.55 & 29.59/31.74/32.22 & 21.57/22.70/22.60   & 21.96/23.39/22.91 \\
                    & P@10 & 21.83            & 39.58             & 27.06/28.99/29.78 & 33.41/36.99/37.23 & 24.34/25.56/26.58   & 24.34/26.49/27.21 \\
                    & P@30 & 26.41            & 45.1              & 33.00/34.65/34.72 & 40.33/42.72/43.68 & 29.96/31.19/30.88   & 31.03/32.22/32.46 \\ \hline
\multirow{4}{*}{fi} & P@1  & 20.54            & 35.15             & 22.91/25.34/25.98 & 27.92/30.79/32.22 & 20.65/23.01/23.31   & 22.20/24.82/24.34 \\
                    & P@5  & 31.64            & 47.39             & 35.86/37.87/38.51 & 43.20/46.30/45.82 & 32.92/34.25/35.38   & 34.13/36.52/37.95 \\
                    & P@10 & 36.65            & 51.61             & 40.23/41.52/42.09 & 48.93/49.88/50.84 & 36.50/37.83/38.45   & 38.19/40.81/41.05 \\
                    & P@30 & 43.95            & 59.7              & 48.03/49.11/49.89 & 58.00/58.95/58.47 & 43.66/44.79/46.22   & 45.58/47.26/48.45 \\ \hline
\multirow{4}{*}{ja} & P@1  & 19.26            & 37.15             & 20.69/21.76/22.69 & 23.15/25.78/24.34 & 19.53/20.04/21.98   & 22.20/22.91/24.82 \\
                    & P@5  & 32.86            & 52.83             & 36.86/37.80/39.80 & 41.53/42.48/44.63 & 34.87/35.69/37.73   & 38.42/38.66/40.10 \\
                    & P@10 & 39.23            & 58.12             & 43.81/44.45/45.81 & 49.40/48.69/50.60 & 41.41/42.84/43.76   & 45.11/45.35/45.82 \\
                    & P@30 & 47.67            & 64.92             & 53.61/54.47/55.69 & 59.19/60.38/61.81 & 51.02/51.94/53.07   & 52.74/53.94/54.89 \\ \hline
\multirow{4}{*}{th} & P@1  & 15.32            & 38.3              & 20.62/22.41/21.62 & 26.73/29.83/28.40 & 17.89/19.22/18.71   & 19.33/21.48/20.05 \\
                    & P@5  & 27.56            & 54.26             & 38.37/40.52/39.94 & 44.39/47.73/48.93 & 35.79/37.32/36.09   & 37.47/39.14/38.66 \\
                    & P@10 & 34.14            & 59.2              & 45.10/47.53/47.10 & 53.46/56.56/56.09 & 41.51/43.76/43.25   & 43.20/44.87/45.35 \\
                    & P@30 & 45.1             & 68.79             & 55.48/56.76/56.62 & 62.29/64.44/64.92 & 52.56/53.48/53.07   & 54.18/54.65/55.37 \\ \hline
\end{tabular}%
}
\caption{Full results for mT0-xl (3.7B) with prompt-based embedding}
\label{tab:my-table}
\end{table*}

\begin{table*}[]
\centering
\resizebox{0.8\textwidth}{!}{%
\begin{tabular}{|c|l|c|c|c|c|c|c|}
\hline
                    & \multicolumn{7}{c|}{\begin{tabular}[c]{@{}c@{}}EN\\ (P@K: 1000/2000/3000) \end{tabular}}                                   \\ \hline
Language            & P@K  & Before-Alignment & Ceiling Performance & Eval - Both       & Eval - Abstract   & Eval-Physical (all) & Eval - Physical (downsample)   \\ \hline
\multirow{4}{*}{fr} & P@1  & 42.09            & 68.65             & 36.22/39.44/40.59 & 42.96/46.30/48.69 & 33.33/36.50/37.12   & 33.65/38.19/38.42 \\
                    & P@5  & 50.54            & 76.23             & 47.67/49.82/51.54 & 55.13/57.04/59.19 & 44.48/46.73/48.26   & 48.69/50.84/52.27 \\
                    & P@10 & 53.76            & 78.6              & 50.68/53.47/54.69 & 57.52/60.38/61.10 & 47.75/50.51/51.94   & 51.31/54.18/56.09 \\
                    & P@30 & 58.2             & 81.96             & 56.48/58.12/58.77 & 62.29/65.16/65.63 & 53.99/55.11/55.83   & 57.52/58.95/59.90 \\ \hline
\multirow{4}{*}{ro} & P@1  & 34.86            & 72.3              & 27.63/31.42/33.21 & 31.03/35.56/37.95 & 26.18/29.65/31.19   & 28.16/30.79/32.94 \\
                    & P@5  & 44.74            & 77.81             & 40.23/43.31/45.60 & 44.87/48.69/50.12 & 38.24/41.00/43.66   & 42.24/45.11/46.78 \\
                    & P@10 & 47.96            & 80.89             & 44.52/47.60/48.89 & 49.40/53.46/55.13 & 42.43/45.09/46.22   & 46.30/47.97/49.16 \\
                    & P@30 & 53.11            & 84.4              & 49.82/52.97/53.40 & 55.37/58.23/58.71 & 47.44/50.72/51.12   & 51.07/52.51/53.22 \\ \hline
\multirow{4}{*}{eu} & P@1  & 24.62            & 71.22             & 15.68/19.76/22.19 & 15.75/23.39/27.68 & 15.64/18.20/19.84   & 16.23/19.09/21.00 \\
                    & P@5  & 32.57            & 77.52             & 27.92/32.00/33.64 & 34.61/40.10/42.72 & 25.05/28.53/29.75   & 27.45/32.46/31.98 \\
                    & P@10 & 36.15            & 79.81             & 32.64/35.36/36.86 & 40.81/44.15/45.82 & 29.14/31.60/33.03   & 32.46/34.13/35.80 \\
                    & P@30 & 40.09            & 83.75             & 38.80/40.87/41.80 & 47.73/49.88/51.31 & 34.97/37.01/37.73   & 38.66/40.57/42.00 \\ \hline
\multirow{4}{*}{fi} & P@1  & 26.27            & 71.37             & 15.89/20.90/22.33 & 17.42/23.87/25.54 & 15.24/19.63/20.96   & 14.56/19.81/21.24 \\
                    & P@5  & 34.5             & 78.31             & 31.07/33.72/35.72 & 33.65/36.99/40.10 & 29.96/32.31/33.84   & 32.70/35.56/37.47 \\
                    & P@10 & 37.51            & 80.17             & 35.58/37.87/39.44 & 39.14/40.33/43.44 & 34.05/36.81/37.73   & 37.23/40.81/41.29 \\
                    & P@30 & 42.73            & 83.75             & 41.73/44.31/45.38 & 46.54/48.21/49.64 & 39.67/42.64/43.56   & 42.48/45.11/45.58 \\ \hline
\multirow{4}{*}{ja} & P@1  & 28.49            & 80.17             & 13.31/20.11/23.26 & 11.22/19.33/24.11 & 14.21/20.45/22.90   & 17.18/22.20/25.30 \\
                    & P@5  & 41.16            & 88.83             & 33.50/40.37/43.38 & 32.46/38.19/40.81 & 33.95/41.31/44.48   & 34.84/46.30/49.64 \\
                    & P@10 & 46.46            & 91.2              & 40.23/45.96/48.39 & 37.71/42.48/46.06 & 41.31/47.44/49.39   & 43.44/52.27/54.18 \\
                    & P@30 & 54.62            & 94.27             & 50.47/54.69/55.98 & 48.69/54.42/55.85 & 51.23/54.81/56.03   & 55.13/59.43/59.19 \\ \hline
\multirow{4}{*}{th} & P@1  & 25.91            & 79.1              & 12.03/16.18/19.90 & 11.22/15.99/20.29 & 12.37/16.26/19.73   & 11.93/18.62/22.20 \\
                    & P@5  & 41.95            & 87.54             & 36.58/40.52/43.02 & 36.28/41.29/44.39 & 36.71/40.18/42.43   & 39.14/43.91/46.06 \\
                    & P@10 & 47.24            & 90.26             & 43.66/47.46/49.75 & 43.68/49.16/51.31 & 43.66/46.73/49.08   & 46.30/49.40/51.31 \\
                    & P@30 & 56.41            & 93.27             & 54.47/57.98/59.70 & 56.80/61.34/61.58 & 53.48/56.54/58.90   & 56.09/58.47/62.29 \\ \hline
\end{tabular}%
}
\caption{Full results for mT0-xxl (13B) with average embedding}
\label{tab:my-table}
\end{table*}

\begin{table*}[]
\centering
\resizebox{0.8\textwidth}{!}{%
\begin{tabular}{|c|l|c|c|c|c|c|c|}
\hline
                    & \multicolumn{7}{c|}{\begin{tabular}[c]{@{}c@{}}EN\\ (P@K: 1000/2000/3000) \end{tabular}}                                   \\ \hline
Language            & P@K  & Before-Alignment & Ceiling Performance & Eval - Both       & Eval - Abstract   & Eval-Physical (all) & Eval - Physical (downsample)   \\ \hline
\multirow{4}{*}{fr} & P@1  & 29.71            & 56.62             & 47.82/49.25/49.96 & 58.00/59.90/59.90 & 43.46/44.68/45.71   & 45.11/45.58/46.78 \\
                    & P@5  & 47.96            & 64.85             & 58.34/59.34/59.77 & 68.26/69.69/69.93 & 54.09/54.91/55.42   & 55.85/57.04/57.76 \\
                    & P@10 & 54.9             & 67.86             & 62.49/62.56/62.63 & 71.12/71.84/71.84 & 58.79/58.59/58.69   & 60.14/60.14/60.14 \\
                    & P@30 & 62.92            & 72.73             & 68.36/68.15/68.36 & 77.09/77.09/77.80 & 64.62/64.31/64.31   & 67.06/67.06/66.83 \\ \hline
\multirow{4}{*}{ro} & P@1  & 9.59             & 53.4              & 38.08/40.59/41.16 & 45.82/48.93/49.88 & 34.76/37.01/37.42   & 38.19/40.81/41.29 \\
                    & P@5  & 22.48            & 63.92             & 51.04/52.40/53.54 & 59.90/61.81/62.77 & 47.24/48.36/49.59   & 50.84/52.27/53.70 \\
                    & P@10 & 31.07            & 67.64             & 55.76/57.84/58.20 & 63.72/65.63/65.87 & 52.35/54.50/54.91   & 55.37/57.76/59.67 \\
                    & P@30 & 45.03            & 72.37             & 63.28/64.07/64.64 & 70.88/71.36/71.84 & 60.02/60.94/61.55   & 63.25/64.44/65.39 \\ \hline
\multirow{4}{*}{eu} & P@1  & 1.93             & 16.68             & 2.79/4.65/5.73    & 4.06/6.92/7.88    & 2.25/3.68/4.81      & 3.34/5.25/5.73    \\
                    & P@5  & 5.73             & 26.63             & 8.09/12.10/13.74  & 11.93/16.23/17.66 & 6.44/10.33/12.07    & 8.59/11.93/13.84  \\
                    & P@10 & 8.66             & 30.78             & 12.46/16.18/18.32 & 16.47/22.20/23.87 & 10.74/13.60/15.95   & 13.13/14.80/16.95 \\
                    & P@30 & 15.1             & 39.08             & 20.76/24.84/26.49 & 28.16/34.37/35.32 & 17.59/20.76/22.70   & 19.33/22.43/24.82 \\ \hline
\multirow{4}{*}{fi} & P@1  & 5.15             & 50.47             & 31.78/33.29/34.93 & 36.52/37.95/38.90 & 29.75/31.29/33.23   & 32.46/35.08/37.23 \\
                    & P@5  & 15.46            & 60.34             & 45.88/47.46/48.82 & 53.94/54.65/56.32 & 42.43/44.38/45.60   & 46.30/47.97/49.16 \\
                    & P@10 & 22.48            & 63.92             & 50.11/52.68/53.61 & 58.95/61.34/61.10 & 46.32/48.98/50.41   & 50.60/52.51/53.94 \\
                    & P@30 & 36.72            & 71.73             & 58.41/59.84/59.99 & 67.30/68.50/68.74 & 54.60/56.13/56.24   & 59.90/59.67/59.90 \\ \hline
\multirow{4}{*}{ja} & P@1  & 6.73             & 48.25             & 31.28/32.36/33.36 & 36.28/36.75/38.42 & 29.14/30.47/31.19   & 32.94/35.08/36.28 \\
                    & P@5  & 20.33            & 60.92             & 48.39/49.46/50.25 & 52.27/53.46/54.18 & 46.73/47.75/48.57   & 51.31/52.51/53.70 \\
                    & P@10 & 28.99            & 65.07             & 53.47/54.97/55.91 & 57.28/58.71/59.90 & 51.84/53.37/54.19   & 55.85/56.80/57.76 \\
                    & P@30 & 43.31            & 71.58             & 61.78/62.71/62.42 & 63.72/65.16/64.44 & 60.94/61.66/61.55   & 63.48/64.68/64.68 \\ \hline
\multirow{4}{*}{th} & P@1  & 6.3              & 48.1              & 26.70/28.85/29.06 & 32.22/32.70/34.37 & 24.34/27.20/26.79   & 25.06/29.36/28.64 \\
                    & P@5  & 16.68            & 61.56             & 43.74/46.03/45.96 & 50.12/51.79/51.31 & 41.00/43.56/43.66   & 43.91/47.02/46.78 \\
                    & P@10 & 25.05            & 67.29             & 50.18/52.83/53.26 & 56.09/57.52/57.52 & 47.65/50.82/51.43   & 50.36/53.22/53.22 \\
                    & P@30 & 38.51            & 75.3              & 61.99/63.14/63.71 & 67.06/68.50/68.74 & 59.82/60.84/61.55   & 60.86/61.10/61.34 \\ \hline
\end{tabular}%
}
\caption{Full results for mT0-xxl (13B) with prompt-based embedding}
\label{tab:my-table}
\end{table*}

\begin{table*}[]
\centering
\resizebox{0.8\textwidth}{!}{%
\begin{tabular}{|c|l|c|c|c|c|c|c|}
\hline
                    & \multicolumn{7}{c|}{\begin{tabular}[c]{@{}c@{}}EN\\ (P@K: 1000/2000/3000) \end{tabular}}                                   \\ \hline
Language            & P@K  & Before-Alignment & Ceiling Performance & Eval - Both       & Eval - Abstract   & Eval-Physical (all) & Eval - Physical (downsample)   \\ \hline
\multirow{4}{*}{fr} & p@1  & 25.41            & 87.62             & 19.18/27.56/30.64 & 22.20/32.22/36.04 & 17.89/25.56/28.32   & 17.66/25.06/28.40 \\
                    & p@5  & 32.28            & 92.13             & 33.00/39.30/42.30 & 37.71/46.78/49.16 & 30.98/36.09/39.37   & 32.94/36.75/40.10 \\
                    & p@10 & 35.65            & 93.92             & 36.72/43.52/45.96 & 42.48/49.64/53.46 & 34.25/40.90/42.74   & 36.04/42.00/44.39 \\
                    & p@30 & 42.95            & 95.71             & 44.45/49.53/51.40 & 51.07/56.56/57.52 & 41.62/46.52/48.77   & 43.44/47.97/50.84 \\ \hline
\multirow{4}{*}{ro} & p@1  & 16.39            & 84.40             & 6.94/12.53/15.53  & 6.44/11.46/15.04  & 7.16/12.99/15.75    & 7.40/13.13/16.47  \\
                    & p@5  & 22.05            & 88.40             & 19.11/23.98/26.84 & 21.24/27.21/30.79 & 18.20/22.60/25.15   & 19.57/25.30/27.68 \\
                    & p@10 & 24.70            & 89.41             & 23.62/27.49/30.21 & 26.97/30.79/34.61 & 22.19/26.07/28.32   & 24.58/29.36/30.79 \\
                    & p@30 & 28.85            & 91.34             & 30.21/34.50/36.29 & 36.52/40.10/41.53 & 27.51/32.11/34.05   & 31.03/34.37/36.52 \\ \hline
\multirow{4}{*}{eu} & p@1  & 10.45            & 80.89             & 1.93/4.08/6.73    & 2.86/4.30/7.40    & 1.53/3.99/6.44      & 0.24/2.15/5.49    \\
                    & p@5  & 13.82            & 87.19             & 9.74/13.39/15.32  & 11.46/13.60/16.23 & 9.00/13.29/14.93    & 7.64/12.17/14.08  \\
                    & p@10 & 15.75            & 88.98             & 12.81/16.61/18.83 & 15.04/17.90/20.76 & 11.86/16.05/18.00   & 11.46/15.27/17.18 \\
                    & p@30 & 18.83            & 91.41             & 17.61/22.41/22.91 & 20.53/23.87/24.11 & 16.36/21.78/22.39   & 16.71/21.96/21.48 \\ \hline
\multirow{4}{*}{fi} & p@1  & 7.73             & 74.37             & 0.93/2.65/4.29    & 0.24/1.91/5.01    & 1.23/2.97/3.99      & 1.43/3.82/4.77    \\
                    & p@5  & 10.67            & 82.18             & 7.37/10.24/12.03  & 9.55/13.13/15.51  & 6.44/9.00/10.53     & 6.68/10.26/12.17  \\
                    & p@10 & 12.60            & 84.32             & 11.31/14.75/16.39 & 13.37/18.62/21.24 & 10.43/13.09/14.31   & 12.17/14.08/15.51 \\
                    & p@30 & 16.96            & 87.40             & 18.11/21.33/23.19 & 24.11/26.25/28.16 & 15.54/19.22/21.06   & 17.42/20.05/22.43 \\ \hline
\multirow{4}{*}{ja} & p@1  & 10.09            & 89.62             & 1.93/3.51/5.37    & 3.10/5.73/7.88    & 1.43/2.56/4.29      & 1.43/2.63/5.01    \\
                    & p@5  & 19.90            & 94.27             & 8.80/14.60/17.04  & 11.69/18.14/20.05 & 7.57/13.09/15.75    & 8.35/14.80/16.23  \\
                    & p@10 & 24.12            & 95.78             & 12.88/19.90/23.48 & 16.23/23.63/29.36 & 11.45/18.30/20.96   & 12.89/20.05/23.15 \\
                    & p@30 & 31.28            & 96.78             & 22.41/29.78/33.64 & 27.92/36.52/38.42 & 20.04/26.89/31.60   & 21.72/29.83/33.65 \\ \hline
\multirow{4}{*}{th} & p@1  & 0.79             & 71.37             & 0.21/0.43/0.29    & 0.24/0.72/0.48    & 0.20/0.31/0.20      & 0.48/0.48/0.24    \\
                    & p@5  & 1.72             & 79.46             & 1.50/2.51/3.44    & 2.39/2.86/4.30    & 1.12/2.35/3.07      & 1.43/3.10/3.10    \\
                    & p@10 & 2.51             & 81.10             & 3.51/5.08/6.59    & 3.82/5.25/8.11    & 3.37/5.01/5.93      & 3.58/5.97/6.44    \\
                    & p@30 & 4.08             & 84.54             & 8.23/10.67/12.31  & 8.83/12.17/14.80  & 7.98/10.02/11.25    & 8.11/10.02/11.46 \\ \hline
\end{tabular}%
}
\caption{Full results for Llama2-7B with last-token embedding}
\label{tab:my-table}
\end{table*}

\begin{table*}[]
\centering
\resizebox{0.8\textwidth}{!}{%
\begin{tabular}{|c|l|c|c|c|c|c|c|}
\hline
                    & \multicolumn{7}{c|}{\begin{tabular}[c]{@{}c@{}}EN\\ (P@K: 1000/2000/3000) \end{tabular}}                                   \\ \hline
Language            & P@K  & Before-Alignment & Ceiling Performance & Eval - Both       & Eval - Abstract   & Eval-Physical (all) & Eval - Physical (downsample)   \\ \hline
\multirow{4}{*}{fr} & p@1  & 60.42            & 86.97             & 42.16/48.03/51.40 & 52.51/57.52/61.58 & 37.73/43.97/47.03   & 35.08/43.91/46.30 \\
                    & p@5  & 72.23            & 93.06             & 66.43/68.29/69.65 & 73.99/75.89/76.13 & 63.19/65.03/66.87   & 65.87/68.26/69.45 \\
                    & p@10 & 75.52            & 95.35             & 72.01/73.94/74.16 & 78.04/79.47/80.19 & 69.43/71.57/71.57   & 73.51/74.46/74.22 \\
                    & p@30 & 79.53            & 98.00             & 78.02/79.03/79.67 & 84.25/84.73/84.96 & 75.36/76.58/77.40   & 78.76/80.67/81.62 \\ \hline
\multirow{4}{*}{ro} & p@1  & 43.02            & 81.53             & 19.83/29.06/32.71 & 26.73/37.71/39.62 & 16.87/25.36/29.75   & 15.75/24.11/28.64 \\
                    & p@5  & 55.05            & 91.91             & 47.60/51.40/52.97 & 57.28/58.95/60.38 & 43.46/48.16/49.80   & 44.63/49.16/51.31 \\
                    & p@10 & 59.20            & 94.77             & 53.04/55.98/56.91 & 61.81/64.44/64.44 & 49.28/52.35/53.68   & 50.84/52.98/55.13 \\
                    & p@30 & 63.71            & 97.64             & 60.70/62.13/63.28 & 68.97/71.12/71.84 & 57.16/58.18/59.61   & 58.23/59.43/59.67 \\ \hline
\multirow{4}{*}{eu} & p@1  & 15.53            & 70.65             & 4.29/8.16/11.60   & 7.16/10.74/15.99  & 3.07/7.06/9.71      & 1.67/6.21/8.83    \\
                    & p@5  & 24.77            & 81.96             & 16.96/22.91/26.34 & 24.58/28.88/32.70 & 13.70/20.35/23.62   & 12.17/20.29/22.43 \\
                    & p@10 & 27.13            & 84.54             & 21.83/27.13/30.35 & 27.92/33.65/36.28 & 19.22/24.34/27.81   & 18.38/25.06/27.68 \\
                    & p@30 & 30.71            & 89.33             & 28.42/32.78/35.08 & 34.13/38.66/41.29 & 25.97/30.27/32.41   & 26.49/31.26/34.13 \\ \hline
\multirow{4}{*}{fi} & p@1  & 25.05            & 75.81             & 7.37/11.31/15.10  & 11.69/15.99/21.72 & 5.52/9.30/12.27     & 5.97/9.07/11.46   \\
                    & p@5  & 35.22            & 88.12             & 26.77/31.50/33.64 & 35.32/38.42/40.33 & 23.11/28.53/30.78   & 25.78/31.26/32.94 \\
                    & p@10 & 38.73            & 91.48             & 33.57/37.15/39.44 & 40.33/44.15/47.02 & 30.67/34.15/36.20   & 34.37/36.04/38.19 \\
                    & p@30 & 44.38            & 95.13             & 41.45/45.53/47.03 & 50.84/55.61/57.52 & 37.53/41.21/42.54   & 38.90/43.44/44.15 \\ \hline
\multirow{4}{*}{ja} & p@1  & 26.91            & 84.18             & 11.31/16.96/21.83 & 14.32/20.05/24.34 & 10.02/15.64/20.76   & 10.74/17.18/22.67 \\
                    & p@5  & 47.75            & 93.84             & 38.58/45.74/49.25 & 47.49/53.70/55.61 & 34.76/42.33/46.52   & 37.23/45.11/48.21 \\
                    & p@10 & 54.90            & 96.06             & 47.67/53.90/56.62 & 56.09/62.05/64.68 & 44.07/50.41/53.17   & 46.78/53.46/56.09 \\
                    & p@30 & 62.13            & 97.85             & 59.48/64.35/67.07 & 67.06/71.84/73.75 & 56.24/61.15/64.21   & 58.95/64.44/67.30 \\ \hline
\multirow{4}{*}{th} & p@1  & 2.58             & 38.15             & 0.57/1.36/1.79    & 0.48/1.19/2.39    & 0.61/1.43/1.53      & 0.72/1.43/1.43    \\
                    & p@5  & 6.59             & 57.19             & 6.23/8.88/9.38    & 8.11/13.37/13.13  & 5.42/6.95/7.77      & 5.25/7.16/7.40    \\
                    & p@10 & 9.74             & 64.21             & 10.81/14.17/14.46 & 13.60/18.14/18.38 & 9.61/12.47/12.78    & 9.31/11.22/11.22  \\
                    & p@30 & 17.04            & 75.52             & 19.83/24.41/25.34 & 24.11/29.12/30.31 & 18.00/22.39/23.21   & 17.90/20.05/21.96 \\ \hline
\end{tabular}%
}
\caption{Full results for Llama2-7B with prompt-based embedding}
\label{tab:my-table}
\end{table*}

\begin{table*}[]
\centering
\resizebox{0.8\textwidth}{!}{%
\begin{tabular}{|c|l|c|c|c|c|c|c|}
\hline
                    & \multicolumn{7}{c|}{\begin{tabular}[c]{@{}c@{}}EN\\ (P@K: 1000/2000/3000) \end{tabular}}                                   \\ \hline
Language            & P@K  & Before-Alignment & Ceiling Performance & Eval - Both       & Eval - Abstract   & Eval-Physical (all) & Eval - Physical (downsample)   \\ \hline
\multirow{4}{*}{fr} & p@1  & 30.78            & 98.50             & 15.39/24.84/32.00 & 19.57/27.92/37.71 & 13.60/23.52/29.55   & 11.93/21.48/28.40 \\
                    & p@5  & 39.73            & 99.14             & 34.72/43.74/47.10 & 40.10/48.45/52.98 & 32.41/41.72/44.58   & 31.98/42.24/45.58 \\
                    & p@10 & 43.38            & 99.28             & 40.23/48.25/51.90 & 44.63/54.65/58.47 & 38.34/45.50/49.08   & 38.66/45.35/51.31 \\
                    & p@30 & 50.82            & 99.64             & 49.25/55.69/57.55 & 53.94/61.10/62.29 & 47.24/53.37/55.52   & 49.64/54.89/57.04 \\ \hline
\multirow{4}{*}{ro} & p@1  & 21.12            & 96.13             & 4.44/10.45/14.60  & 3.58/10.26/14.32  & 4.81/10.53/14.72    & 4.30/9.79/14.56   \\
                    & p@5  & 26.70            & 97.71             & 18.68/25.27/28.06 & 21.72/29.36/32.22 & 17.38/23.52/26.28   & 17.90/24.82/27.92 \\
                    & p@10 & 30.64            & 97.92             & 24.05/30.92/32.86 & 28.88/36.04/36.52 & 21.98/28.73/31.29   & 23.63/30.07/32.22 \\
                    & p@30 & 35.86            & 98.28             & 32.93/39.08/40.87 & 37.71/42.72/45.82 & 30.88/37.53/38.75   & 33.41/40.81/42.72 \\ \hline
\multirow{4}{*}{eu} & p@1  & 13.67            & 96.99             & 1.22/3.58/5.65    & 0.95/3.10/5.49    & 1.33/3.78/5.73      & 0.95/3.34/5.73    \\
                    & p@5  & 18.90            & 98.28             & 9.31/14.46/16.46  & 10.74/15.27/17.90 & 8.69/14.11/15.85    & 8.59/14.08/14.56  \\
                    & p@10 & 20.97            & 99.07             & 13.39/18.61/20.97 & 16.47/21.24/23.87 & 12.07/17.48/19.73   & 11.69/16.95/19.57 \\
                    & p@30 & 24.41            & 99.43             & 19.90/25.84/27.63 & 22.91/29.59/31.26 & 18.61/24.23/26.07   & 18.38/25.54/26.73 \\ \hline
\multirow{4}{*}{fi} & p@1  & 9.52             & 93.70             & 0.72/2.65/3.94    & 0.24/1.91/3.10    & 0.92/2.97/4.29      & 0.48/2.39/4.30    \\
                    & p@5  & 14.10            & 96.49             & 7.73/11.31/13.82  & 8.83/12.65/15.75  & 7.26/10.74/12.99    & 7.40/10.98/13.84  \\
                    & p@10 & 17.32            & 97.28             & 11.17/15.82/18.75 & 11.46/18.62/22.67 & 11.04/14.62/17.08   & 12.41/16.23/18.14 \\
                    & p@30 & 22.48            & 98.14             & 20.54/25.63/27.99 & 24.82/30.31/32.22 & 18.71/23.62/26.18   & 20.29/25.30/27.92 \\ \hline
\multirow{4}{*}{ja} & p@1  & 24.19            & 98.14             & 1.65/4.58/6.37    & 1.19/5.49/7.88    & 1.84/4.19/5.73      & 2.39/4.77/4.77    \\
                    & p@5  & 33.07            & 98.78             & 12.53/19.54/23.84 & 16.47/22.67/28.88 & 10.84/18.20/21.68   & 10.98/20.53/22.43 \\
                    & p@10 & 38.01            & 99.28             & 18.61/24.98/30.85 & 22.91/27.21/33.65 & 16.77/24.03/29.65   & 17.42/26.49/31.74 \\
                    & p@30 & 46.46            & 99.43             & 27.63/36.79/40.23 & 30.31/39.62/43.91 & 26.48/35.58/38.65   & 28.64/37.47/40.81 \\ \hline
\multirow{4}{*}{th} & p@1  & 0.86             & 90.19             & 0.00/0.21/0.36    & 0.00/0.72/1.19    & 0.00/0.00/0.00      & 0.00/0.00/0.00    \\
                    & p@5  & 1.43             & 95.06             & 3.01/4.29/6.37    & 2.86/4.06/5.97    & 3.07/4.40/6.54      & 3.34/4.77/7.88    \\
                    & p@10 & 2.08             & 95.71             & 5.37/7.23/9.59    & 5.49/6.92/9.31    & 5.32/7.36/9.71      & 6.44/8.11/10.26   \\
                    & p@30 & 3.29             & 97.14             & 10.16/13.53/16.54 & 11.46/13.60/17.66 & 9.61/13.50/16.05    & 10.02/13.84/16.23 \\ \hline
\end{tabular}%
}
\caption{Full results for Llama2-13B with last-token embedding}
\label{tab:my-table}
\end{table*}

\begin{table*}[]
\centering
\resizebox{0.8\textwidth}{!}{%
\begin{tabular}{|c|l|c|c|c|c|c|c|}
\hline
                    & \multicolumn{7}{c|}{\begin{tabular}[c]{@{}c@{}}EN\\ (P@K: 1000/2000/3000) \end{tabular}}                                   \\ \hline
Language            & P@K  & Before-Alignment & Ceiling Performance & Eval - Both       & Eval - Abstract   & Eval-Physical (all) & Eval - Physical (downsample)   \\ \hline
\multirow{4}{*}{fr} & p@1  & 59.27            & 87.47             & 44.95/50.68/53.61 & 52.74/60.62/63.48 & 41.62/46.42/49.39   & 41.29/47.49/50.12 \\
                    & p@5  & 72.01            & 95.71             & 68.50/71.08/71.87 & 77.09/78.04/79.00 & 64.83/68.10/68.81   & 67.06/70.17/69.45 \\
                    & p@10 & 74.95            & 97.35             & 72.58/74.30/75.38 & 79.95/81.38/82.82 & 69.43/71.27/72.19   & 71.12/72.55/73.51 \\
                    & p@30 & 79.24            & 98.35             & 78.45/79.17/79.53 & 84.25/84.96/84.96 & 75.97/76.69/77.20   & 77.33/78.04/79.24 \\ \hline
\multirow{4}{*}{ro} & p@1  & 38.87            & 82.39             & 24.98/34.65/36.65 & 31.50/44.39/46.06 & 22.19/30.47/32.62   & 22.91/29.12/33.41 \\
                    & p@5  & 52.97            & 93.41             & 51.40/54.76/56.76 & 60.38/63.25/63.96 & 47.55/51.12/53.68   & 49.40/52.51/54.65 \\
                    & p@10 & 56.55            & 95.35             & 56.41/58.98/60.34 & 64.92/67.30/67.54 & 52.76/55.42/57.26   & 55.61/55.85/58.23 \\
                    & p@30 & 62.99            & 97.35             & 63.06/65.50/66.14 & 70.17/73.03/73.27 & 60.02/62.27/63.09   & 62.29/63.72/64.68 \\ \hline 
\multirow{4}{*}{eu} & p@1  & 20.26            & 75.30             & 6.08/10.95/15.75  & 6.21/11.22/17.42  & 6.03/10.84/15.03    & 5.73/10.74/14.08  \\
                    & p@5  & 29.49            & 86.83             & 20.40/28.06/31.78 & 23.15/31.26/33.89 & 19.22/26.69/30.88   & 17.90/26.25/29.59 \\
                    & p@10 & 31.64            & 89.84             & 24.41/31.93/35.15 & 27.21/34.84/36.75 & 23.21/30.67/34.46   & 22.67/30.55/33.89 \\
                    & p@30 & 35.72            & 92.84             & 30.42/37.37/39.44 & 31.74/38.66/40.33 & 29.86/36.81/39.06   & 30.55/38.19/39.14 \\ \hline
\multirow{4}{*}{fi} & p@1  & 23.91            & 79.96             & 9.81/15.03/18.68  & 12.65/17.66/21.00 & 8.59/13.91/17.69    & 7.88/14.56/18.62  \\
                    & p@5  & 33.86            & 90.91             & 30.06/34.22/37.01 & 33.41/39.86/40.81 & 28.63/31.80/35.48   & 28.64/32.70/36.99 \\
                    & p@10 & 36.79            & 92.84             & 34.86/39.66/41.88 & 40.81/45.58/48.45 & 32.31/37.12/39.06   & 33.89/36.52/39.62 \\
                    & p@30 & 41.45            & 95.85             & 41.80/46.39/47.89 & 49.16/53.46/55.13 & 38.65/43.35/44.79   & 39.62/43.20/45.35 \\ \hline
\multirow{4}{*}{ja} & p@1  & 15.60            & 88.05             & 12.74/19.69/23.48 & 13.13/20.76/26.01 & 12.58/19.22/22.39   & 12.65/19.81/23.39 \\
                    & p@5  & 28.35            & 96.99             & 43.38/51.40/53.47 & 47.49/58.71/60.62 & 41.62/48.26/50.41   & 43.91/51.07/54.42 \\
                    & p@10 & 36.29            & 98.21             & 51.61/58.34/61.42 & 57.52/65.87/68.26 & 49.08/55.11/58.49   & 52.03/58.00/61.10 \\
                    & p@30 & 47.60            & 99.28             & 62.71/68.29/70.29 & 70.41/76.61/77.09 & 59.41/64.72/67.38   & 63.01/66.59/69.45 \\ \hline
\multirow{4}{*}{th} & p@1  & 1.36             & 46.39             & 0.79/1.57/1.93    & 1.43/1.43/2.15    & 0.51/1.64/1.84      & 0.24/1.91/1.91    \\
                    & p@5  & 3.51             & 63.21             & 5.65/7.73/9.52    & 7.40/10.74/12.65  & 4.91/6.44/8.18      & 4.77/5.25/7.40    \\
                    & p@10 & 5.44             & 70.01             & 8.88/12.53/13.60  & 10.26/15.99/16.95 & 8.28/11.04/12.17    & 7.88/8.83/10.74   \\
                    & p@30 & 8.52             & 77.88             & 15.75/19.40/21.83 & 18.14/21.00/24.34 & 14.72/18.71/20.76   & 14.56/17.90/19.09 \\ \hline
\end{tabular}%
}
\caption{Full results for Llama2-13B with prompt-based embedding}
\label{tab:my-table} 
\end{table*}

\begin{table*}[]
\centering
\resizebox{0.8\textwidth}{!}{%
\begin{tabular}{|c|l|c|c|c|c|c|c|}
\hline
                    & \multicolumn{7}{c|}{\begin{tabular}[c]{@{}c@{}}EN\\ (P@K: 1000/2000/3000) \end{tabular}}                                   \\ \hline
Language            & P@K  & Before-Alignment & Ceiling Performance & Eval - Both       & Eval - Abstract   & Eval-Physical (all) & Eval - Physical (downsample)   \\ \hline
\multirow{4}{*}{fr} & p@1  & 19.97            & 85.90             & 13.17/20.04/24.41 & 14.08/22.20/25.54 & 12.78/19.12/23.93   & 11.22/17.66/23.87 \\
                    & p@5  & 23.62            & 89.33             & 27.06/34.79/38.30 & 27.92/38.19/42.24 & 26.69/33.33/36.61   & 26.97/33.41/36.75 \\
                    & p@10 & 26.63            & 90.41             & 32.07/39.58/42.59 & 35.80/44.39/46.78 & 30.47/37.53/40.80   & 31.03/36.99/42.48 \\
                    & p@30 & 30.57            & 92.13             & 39.44/45.88/48.32 & 44.39/51.55/52.98 & 37.32/43.46/46.32   & 38.42/44.39/48.45 \\ \hline
\multirow{4}{*}{ro} & p@1  & 15.18            & 81.39             & 4.51/7.52/10.52   & 3.34/6.44/10.26   & 5.01/7.98/10.63     & 3.82/7.64/10.98   \\
                    & p@5  & 17.47            & 85.25             & 14.39/18.90/21.55 & 15.04/19.09/21.72 & 14.11/18.81/21.47   & 15.04/19.81/23.15 \\
                    & p@10 & 18.97            & 87.33             & 18.90/24.27/26.63 & 20.29/25.78/28.88 & 18.30/23.62/25.66   & 19.09/24.34/27.21 \\
                    & p@30 & 22.55            & 90.12             & 26.77/31.21/34.72 & 29.83/33.89/39.38 & 25.46/30.06/32.72   & 26.25/31.98/34.61 \\ \hline
\multirow{4}{*}{eu} & p@1  & 8.02             & 87.97             & 1.36/2.43/4.29    & 1.91/2.86/3.58    & 1.12/2.25/4.60      & 0.72/2.39/4.77    \\
                    & p@5  & 9.81             & 91.62             & 7.37/11.02/13.17  & 8.35/11.93/15.04  & 6.95/10.63/12.37    & 6.92/11.93/12.89  \\
                    & p@10 & 11.60            & 92.63             & 10.16/14.53/16.39 & 11.69/16.95/18.62 & 9.51/13.50/15.44    & 9.55/15.04/15.99  \\
                    & p@30 & 14.46            & 94.42             & 16.46/20.69/22.41 & 19.57/23.87/25.54 & 15.13/19.33/21.06   & 14.32/20.76/22.20 \\ \hline
\multirow{4}{*}{fi} & p@1  & 5.65             & 79.96             & 1.57/2.51/4.01    & 2.63/2.86/4.53    & 1.12/2.35/3.78      & 0.72/0.95/2.63    \\
                    & p@5  & 8.45             & 85.18             & 7.52/10.74/13.17  & 10.02/12.89/15.75 & 6.44/9.82/12.07     & 5.97/10.26/12.41  \\
                    & p@10 & 10.02            & 86.61             & 12.31/15.25/18.04 & 15.99/17.90/20.76 & 10.74/14.11/16.87   & 10.50/15.27/18.62 \\
                    & p@30 & 14.53            & 88.83             & 19.76/22.69/24.84 & 24.58/25.78/28.88 & 17.69/21.37/23.11   & 17.90/22.67/23.87 \\ \hline
\multirow{4}{*}{ja} & p@1  & 11.81            & 95.13             & 3.15/6.16/8.52    & 3.82/7.64/10.02   & 2.86/5.52/7.87      & 3.58/7.16/9.31    \\
                    & p@5  & 22.05            & 96.99             & 12.46/19.33/23.41 & 12.65/20.29/25.54 & 12.37/18.92/22.49   & 13.37/20.53/23.39 \\
                    & p@10 & 25.70            & 97.28             & 18.25/25.48/29.63 & 19.09/26.97/31.98 & 17.89/24.85/28.63   & 20.29/26.25/29.36 \\
                    & p@30 & 31.78            & 98.35             & 27.63/35.43/39.58 & 30.31/37.23/40.33 & 26.48/34.66/39.26   & 28.16/36.28/40.81 \\ \hline
\multirow{4}{*}{th} & p@1  & 0.64             & 89.91             & 0.21/0.43/0.64    & 0.24/0.24/0.48    & 0.20/0.51/0.72      & 0.24/1.19/1.19    \\
                    & p@5  & 1.57             & 92.98             & 2.72/4.58/6.23    & 2.39/4.77/6.92    & 2.86/4.50/5.93      & 3.58/5.25/6.68    \\
                    & p@10 & 2.58             & 93.49             & 4.72/6.30/8.95    & 5.49/6.68/10.74   & 4.40/6.13/8.18      & 4.53/6.44/8.35    \\
                    & p@30 & 4.37             & 94.56             & 10.31/13.53/15.39 & 11.93/15.99/17.18 & 9.61/12.47/14.62    & 9.55/12.89/14.80 \\ \hline
\end{tabular}%
}
\caption{Full results for Llama2-70B with last-token embedding}
\label{tab:my-table}
\end{table*}

\begin{table*}[]
\centering
\resizebox{0.8\textwidth}{!}{%
\begin{tabular}{|c|l|c|c|c|c|c|c|}
\hline
                    & \multicolumn{7}{c|}{\begin{tabular}[c]{@{}c@{}}EN\\ (P@K: 1000/2000/3000) \end{tabular}}                                   \\ \hline
Language            & P@K  & Before-Alignment & Ceiling Performance & Eval - Both       & Eval - Abstract   & Eval-Physical (all) & Eval - Physical (downsample)   \\ \hline
\multirow{4}{*}{fr} & p@1  & 61.92            & 96.06             & 34.00/44.31/48.96 & 42.00/53.22/56.56 & 30.57/40.49/45.71   & 30.31/41.29/45.35 \\
                    & p@5  & 72.15            & 99.36             & 66.43/69.86/71.44 & 74.22/77.80/79.47 & 63.09/66.46/68.00   & 65.63/68.26/69.69 \\
                    & p@10 & 74.66            & 99.50             & 71.73/74.16/75.02 & 78.52/81.38/81.38 & 68.81/71.06/72.29   & 71.84/73.03/74.22 \\
                    & p@30 & 78.53            & 99.93             & 77.17/78.95/79.96 & 83.05/84.25/84.49 & 74.64/76.69/78.02   & 77.57/79.00/80.19 \\ \hline
\multirow{4}{*}{ro} & p@1  & 45.67            & 92.20             & 16.96/26.56/30.49 & 22.43/36.52/37.95 & 14.62/22.29/27.30   & 13.84/21.48/26.01 \\
                    & p@5  & 55.19            & 97.71             & 47.96/53.97/55.33 & 55.13/60.62/61.81 & 44.89/51.12/52.56   & 45.35/51.07/52.98 \\
                    & p@10 & 57.84            & 98.57             & 54.76/58.91/60.63 & 61.58/66.11/68.74 & 51.84/55.83/57.16   & 52.74/55.85/57.04 \\
                    & p@30 & 61.27            & 99.50             & 63.42/65.43/66.86 & 73.27/73.51/74.46 & 59.20/61.96/63.60   & 59.90/63.25/65.16 \\ \hline
\multirow{4}{*}{eu} & p@1  & 20.19            & 87.54             & 3.65/6.87/10.09   & 7.16/10.50/13.84  & 2.15/5.32/8.49      & 2.86/6.92/9.79    \\
                    & p@5  & 27.49            & 95.28             & 19.76/26.34/29.13 & 23.63/31.50/33.41 & 18.10/24.13/27.30   & 19.57/25.06/27.45 \\
                    & p@10 & 29.78            & 96.85             & 24.34/30.06/33.21 & 26.73/34.61/36.52 & 23.31/28.12/31.80   & 25.54/29.12/32.94 \\
                    & p@30 & 31.93            & 98.14             & 30.49/36.08/38.58 & 33.65/41.77/42.96 & 29.14/33.64/36.71   & 30.55/35.08/38.19 \\ \hline
\multirow{4}{*}{fi} & p@1  & 22.91            & 85.61             & 6.01/10.52/13.46  & 9.07/15.99/21.00  & 4.70/8.18/10.22     & 4.53/8.59/10.26   \\
                    & p@5  & 29.21            & 92.70             & 27.77/33.86/36.36 & 35.32/42.00/43.91 & 24.54/30.37/33.13   & 26.01/31.26/34.13 \\
                    & p@10 & 31.85            & 94.56             & 33.50/39.16/41.59 & 41.77/47.26/49.40 & 29.96/35.69/38.24   & 32.22/37.23/39.62 \\
                    & p@30 & 34.72            & 96.49             & 42.09/45.74/47.17 & 49.40/53.70/54.65 & 38.96/42.33/43.97   & 42.24/44.87/45.82 \\ \hline
\multirow{4}{*}{ja} & p@1  & 29.42            & 97.21             & 12.67/18.18/21.83 & 14.56/21.00/23.87 & 11.86/16.97/20.96   & 13.13/17.90/21.72 \\
                    & p@5  & 44.74            & 99.36             & 44.02/51.83/55.69 & 52.51/58.71/62.77 & 40.39/48.88/52.66   & 44.87/53.94/57.28 \\
                    & p@10 & 51.11            & 99.43             & 53.33/60.77/63.85 & 61.34/68.02/71.60 & 49.90/57.67/60.53   & 53.70/61.34/65.16 \\
                    & p@30 & 60.49            & 99.86             & 65.00/70.87/73.30 & 72.55/78.28/80.19 & 61.76/67.69/70.35   & 64.68/69.69/72.79 \\ \hline
\multirow{4}{*}{th} & p@1  & 2.43             & 58.98             & 1.07/1.43/2.15    & 0.48/1.19/2.63    & 1.33/1.53/1.94      & 1.67/1.67/1.91    \\
                    & p@5  & 5.44             & 74.80             & 7.16/9.88/11.52   & 8.59/11.69/14.56  & 6.54/9.10/10.22     & 7.16/8.59/9.31    \\
                    & p@10 & 7.66             & 79.24             & 11.81/16.25/17.32 & 13.60/20.05/20.76 & 11.04/14.62/15.85   & 11.22/14.80/15.51 \\
                    & p@30 & 12.03            & 86.54             & 20.69/25.05/27.27 & 26.25/30.31/30.55 & 18.30/22.80/25.87   & 16.95/21.72/24.34 \\ \hline
\end{tabular}%
}
\caption{Full results for Llama2-70B with prompt-based embedding}
\label{tab:my-table} 
\end{table*}

\end{document}